\documentclass[10pt,twocolumn,letterpaper]{article}

\usepackage[pagenumbers]{iccv} % To force page numbers, e.g. for an arXiv version

\definecolor{iccvblue}{rgb}{0.21,0.49,0.74}
\usepackage[pagebackref,breaklinks,colorlinks,allcolors=iccvblue]{hyperref}

\def\paperID{10548} % *** Enter the Paper ID here
\def\confName{ICCV}
\def\confYear{2025}

\usepackage{pifont}
\usepackage{xspace}
\makeatletter
\DeclareRobustCommand\onedot{\futurelet\@let@token\@onedot}
\def\@onedot{\ifx\@let@token.\else.\null\fi\xspace}

\def\ie{\emph{i.e}\onedot}

\makeatother

\DeclareMathOperator*{\argmin}{arg\,min}
\usepackage{wrapfig}

\newcommand{\reqref}[1]{Eq.~(\ref{#1})}

\title{Visibility-Uncertainty-guided 3D Gaussian Inpainting via Scene Conceptional Learning}

\author{
Mingxuan Cui\\
Zhejiang University\\
{\tt\small 22251322@zju.edu.cn}
\and
{Qing Guo}*\\
CFAR and IHPC, A*STAR, Singapore\\
{\tt\small guo\_qing@cfar.a-star.edu.sg}
\and
Yuyi Wang\\
CRRC Zhuzhou Institute \& Tengen Intelligence Institute\\
{\tt\small yuyiwang920@gmail.com}
\and
Hongkai Yu\\
Cleveland State University\\
{\tt\small h.yu19@csuohio.edu}
\and
Di Lin\\
Tianjin University\\
{\tt\small Ande.lin1988@gmail.com}
\and
Qin Zou\\
Wuhan University\\
{\tt\small qinnzou@gmail.com}
\and
Ming-Ming Cheng\\
Nankai University\\
{\tt\small cmm@nankai.edu.cn}
\and
Xi Li\\
Zhejiang University\\
{\tt\small xilizju@zju.edu.cn}
}

\begin{document}
\maketitle
\begin{abstract}
3D Gaussian Splatting (3DGS) has emerged as a powerful and efficient 3D representation for novel view synthesis. This paper extends 3DGS capabilities to inpainting, where masked objects in a scene are replaced with new contents that blend seamlessly with the surroundings. Unlike 2D image inpainting, 3D Gaussian inpainting (3DGI) is challenging in effectively leveraging complementary visual and semantic cues from multiple input views, as occluded areas in one view may be visible in others.
To address this, we propose a method that measures the visibility uncertainties of 3D points across different input views and uses them to guide 3DGI in utilizing complementary visual cues. We also employ uncertainties to learn a semantic concept of scene without the masked object and use a diffusion model to fill masked objects in input images based on the learned concept.
Finally, we build a novel 3DGI framework, VISTA, by integrating VISibility-uncerTainty-guided 3DGI with scene conceptuAl learning. VISTA generates high-quality 3DGS models capable of synthesizing artifact-free and naturally inpainted novel views. Furthermore, our approach extends to handling dynamic distractors arising from temporal object changes, enhancing its versatility in diverse scene reconstruction scenarios.
We demonstrate the superior performance of our method over state-of-the-art techniques using two challenging datasets: the SPIn-NeRF dataset, featuring 10 diverse static 3D inpainting scenes, and an underwater 3D inpainting dataset derived from UTB180, including fast-moving fish as inpainting targets.
\end{abstract}

\section{Introduction}
\label{sec:intro}

\begin{figure*}
\centering
\vspace{-10pt}
\includegraphics[width=0.88\linewidth]{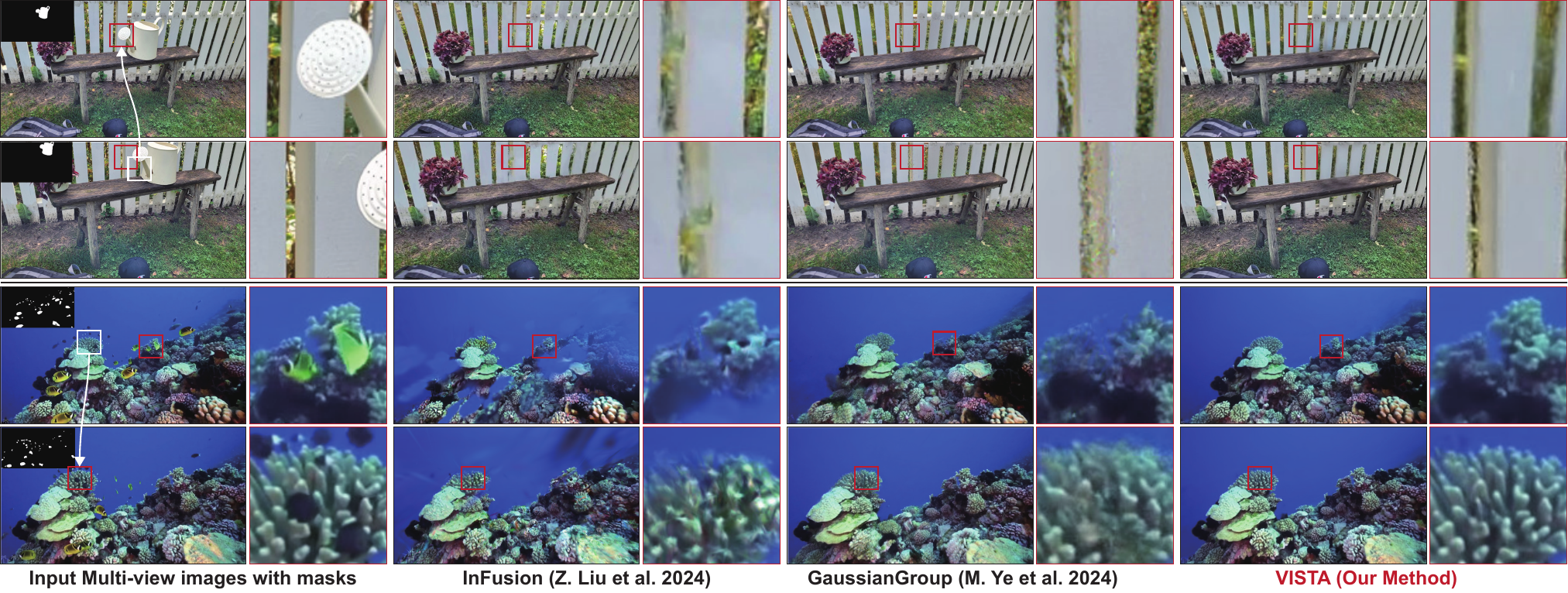}\\
\caption{
Two examples demonstrating the application of two state-of-the-art methods, namely InFusion \citep{liu2024infusion} and GaussianGroup \citep{ye2023gaussian}, alongside our proposed method for 3D Gaussian inpainting to fill masked static and dynamic objects, respectively. The red boxes highlight the advantages of our method and are enlarged on the right side of each image for better visibility. The white boxes and arrows indicate complementary visual cues between two different viewpoints
} 
\label{fig:fig1}
\vspace{-15px}
\end{figure*}

3D representation effectively models a scene and has the ability to synthesize new views of the scene \citep{barron2021mip, mildenhall2021nerf,wang2021neus,kerbl20233d}.
3D Gaussian splatting (3DGS) methods have been demonstrated as efficient and effective ways to represent the scene from a set of images taken from different viewpoints \citep{kerbl20233d,tang2023dreamgaussian,wu20244d}.
Further, enabling editability of 3D scene representations is a cornerstone of technologies like augmented reality and virtual reality \cite{tewari2022advances}. 
3D Gaussian inpainting task is one of the key editing techniques, aiming to replace specified objects with new contents that blend seamlessly with the surroundings.
This capability allows us to: \textit{(1) Remove objects from static scenes:} given multi-view images, we can create a 3D representation that generates novel views with specific objects removed and believably filled in (\Cref{fig:fig1} (Upper)).
\textit{(2) Clean up dynamic scenes:} for scenes with moving elements like fish in water (see \Cref{fig:fig1} (Bottom)), we can build a 3D representation excluding these transient objects, enabling clear, consistent novel view synthesis.

%--------------------------------------------------------------------------------
%--------------------------------------------------------------------------------

However, such an important task is non-trivial and the key challenge is how to leverage the complementary visual and semantic cues from multiple input views.
Intuitively, for a synthesized view, the ideal approach is to replace the targeted erasure region with the occluded content, which naturally completes the inpainting. The key information for this process lies within the other view images, where the obscured areas may be visible from different angles. 
However, how to utilize multi-view information effectively is still an open question.
State-of-the-art works first remove the targeted erasure region-related Gaussians and fill the regions via 2D image inpainting method \citep{ye2023gaussian,wang2024gaussianeditor}, which, however, neglects the complementary cues from other views.
The latest work \citep{liu2024infusion} leverages depth maps of different views to involve the cross-view complementary cues implicitly.
However, depth maps cannot fully represent complementary cues, such as the texture pattern from adjacent perspectives, and the depth project can hardly get high-quality depth maps when moving objects across different views.
As the two cases shown in \Cref{fig:fig1}, InFusion synthesizes new views with obvious artifacts.

In this work, we propose VISibility-uncerTainty-guided 3DGI via scene conceptuAl Learning (VISTA), a novel framework for 3D Gaussian inpainting that leverages complementary visual and semantic cues. Our approach begins by measuring the visibility of 3D points across different views to generate visibility uncertainty maps for each input image. These maps indicate which pixels are most valuable for the inpainting task, based on the principle that pixels visible and consistent from multiple views contribute more significantly.
We then integrate these visibility uncertainty maps into the 3D Gaussian splatting (3DGS) process. This enables the resulting Gaussian model to synthesize new views where masked regions are seamlessly filled with visual information from complementary perspectives.
To address scenarios where large masked regions lack complementary visual cues from other views, we propose learning the concept of the scene without the masked objects. This conceptual learning is guided by the prior inpainting mask and the visibility uncertainty maps derived from the input multi-view images.
The learned concept is then utilized to refine the input images, effectively filling the masked objects through a pre-trained Diffusion model. Furthermore, we implement an iterative process alternating between visibility-uncertainty-guided 3DGI and scene conceptual learning, progressively refining the 3D representation.
As illustrated in \Cref{fig:fig1} (Upper), our method successfully reconstructs high-quality 3D representations of static scenes, naturally filling masked object regions with contextually appropriate content. Additionally, VISTA demonstrates its versatility by effectively removing distractors in dynamic scenes (see \Cref{fig:fig1} (Bottom) for examples).

We demonstrate the superior performance of our method over state-of-the-art techniques using two challenging datasets: the SPIn-NeRF dataset, featuring 10 diverse static 3D in-painting scenes, and an underwater 3D inpainting dataset derived from UTB180, which includes fast-moving fish as inpainting targets. In summary, the contributions of our work are as follows:

\begin{enumerate}
    \item We propose VISibility-uncerTAinty-guided 3D Gaussian inpainting (VISTA-GI) that explicitly leverages multi-view information through visibility uncertainty, achieving 3D Gaussian inpainting for more coherent and accurate scene completions.
    
    \item We propose VISibility-uncerTAinty-guided scene conceptual learning (VISTA-CL) and leverage it for diffusion-based inpainting. VISTA-CL fills masked regions in input images using learned scene concepts, addressing the inpainting task at its core. This approach enhances the fundamental understanding of the scene, leading to more accurate and contextually appropriate inpainting results.
    
    \item We introduce VISTA (VISibility-uncerTainty-guided 3D gaussian inpainTing via scene conceptuAl learning), a novel framework that iteratively combines VISTA-GI and VISTA-CL. This approach simultaneously leverages complementary visual and semantic cues, enhancing 3D Gaussian inpainting with geometric and conceptual information.

    \item We extend VISTA to handle dynamic distractor removal in 3D Gaussian splatting, significantly improving its performance on scenes with temporal variations and outperforming state-of-the-art methods.
\end{enumerate}

%%--------------------------------------------------------------------------------------------
\section{Related Work}
\label{sec:relatedwork}

\subsection{NeRF and 3D Gaussian Splatting}

The challenge of reconstructing a scene from 2D images to obtain suitable new viewpoints is a complex and worthy topic of exploration in computer vision and computer graphics \citep{lombardi2019neural, kutulakos2000theory}. Recently, NeRF \citep{mildenhall2021nerf}  and 3DGS \citep{kerbl20233d} have emerged as two distinct approaches to 3D reconstruction, continuously improving the quality of the reconstructions.

Neural Radiance Fields (NeRF) is an implicit representation method for 3D reconstruction.
It utilizes deep learning techniques to extract the geometric shapes and texture information of objects from images taken from multiple viewpoints, and it uses this information to generate a continuous 3D radiance field, allowing for highly realistic 3D models to be presented from any angle and distance \citep{barron2021mip}. However, their excessively high training and rendering costs \citep{barron2022mip,barron2023zip} often result in poor performance in practical applications.
To resolve these issues, 3D Gaussian splitting (3DGS) is promoted as an explicit representation method that achieves state-of-the-art real-time rendering of high-quality images \citep{lu2024scaffold}. 3DGS explicitly models the space as multiple Gaussian blobs, each with specific 3D positions, opacity, anisotropic covariance, and color features. Through training, it achieves an explicit representation of the three-dimensional space, enabling real-time synthesis of high-quality viewpoint images.

\subsection{2D inpainting and 3D inpainting}
2D inpainting is an elemental task in image generation. The task aims to use the pre-generated mask to create appropriate content for the masked area. Traditional patch-based methods \cite{ruvzic2014context} and later GAN-based \citep{goodfellow2014generative} methods \cite{yu2018generative} could somewhat inpaint regular and small mask areas, but they fail in complex scenes or when there are significant content omissions.
Recently, diffusion models \cite{ho2020denoising,sohl2015deep,song2020score} have become the most powerful technology in inpainting \citep{lugmayr2022repaint,suvorov2022resolution,li2022mat} for their ability to generate new, semantically plausible content. 

Meanwhile, 3D inpainting to edit the scene reconstructed by NeRF or 3DGS is still a challenging task because of the complexity of spatial representation. NeRF-based inpainting \cite{liu2022nerf, mirzaei2023spin, weder2023removing} succeed in inpainting the static objects in the implicit representation. However, their performance is limited because of NeRF's obstacles.  3DGS-based inpainting methods such as Gaussian Grouping \citep{ye2023gaussian}, InFusion \citep{liu2024infusion}, and GaussianEditor \citep{wang2024gaussianeditor} focus on inpainting an existing static Gaussian Splatting scene, but neglecting the dynamic distractors that may appear before obtaining the static scene. 
GScream \citep{wang2025learning} focuses on removing objects by introducing monocular depth estimation and employing cross-attention to enhance texture. It remains a method focused on static objects.
SpotLessSplats \citep{sabourgoli2024spotlesssplats} notices the dynamic distractors and repairs these areas using the pre-predicted masks, but it fails to repair occluded and completely unseen areas.
%%--------------------------------------------------------------------------------------------

\begin{figure*}[t]
\centering
\vspace{-10pt}
\includegraphics[width=0.88\linewidth]{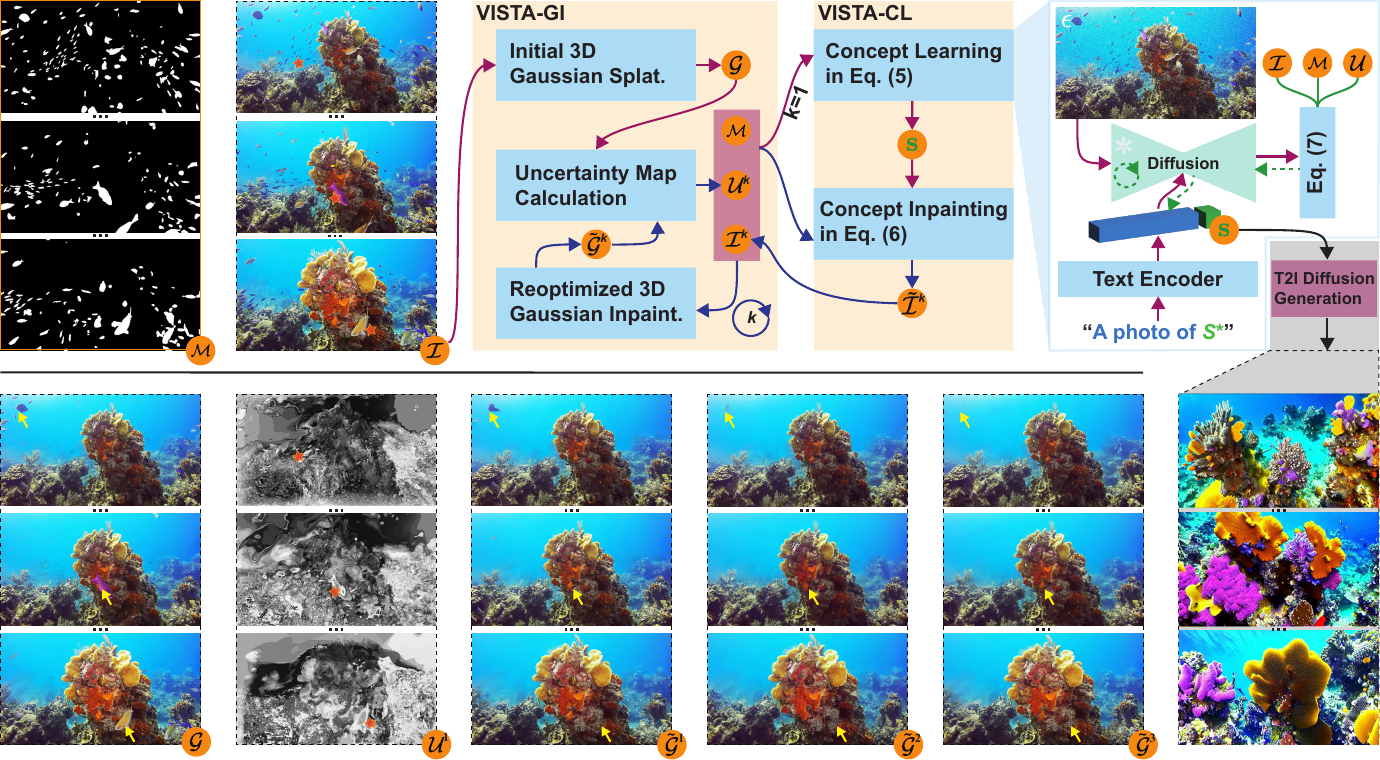}\\
\caption{ 
The f ramework of VISTA comprises two modules: VISTA-GI (described in \Cref{subsec:vista-gi}) and VISTA-CL (detailed in \Cref{subsec:vista-cl}). Results from three views are displayed for key variables in the framework. Note that $\mathcal{G}$, $\tilde{\mathcal{G}}^1$, $\tilde{\mathcal{G}}^2$, and $\tilde{\mathcal{G}}^3$ are 3DGS representations, and the displayed examples are rendered from these representations. The last column shows generated images derived from the learned scene concept. In the uncertainty map, we use \ding{75} to highlight areas of high uncertainty, which denote points (e.g., dynamic fishes) visible from only a few views. Yellow arrows demonstrate the progressive improvement in inpainting quality achieved by our method.
} 
\label{fig:pipeline}
\vspace{-15px}
\end{figure*}

\section{Preliminaries: 3D Gaussian Splatting and Inpainting}
\label{sec:preliminaries}

\subsection{3D Gaussian Splatting}
\label{subsec:3dgs}

Given a set of images $\mathcal{I} = \{\mathbf{I}_i\}_{i=1}^N$ captured from various viewpoints and timestamps, 3D Gaussian splatting (3DGS) aims to learn a collection of anisotropic Gaussian splats $\mathcal{G}=\{\mathbf{g}_j\}_{j=1}^M$ from these multi-view images. Each splat $\mathbf{g}_j$ is characterized by a Gaussian function with mean $\mu_j$, a positive semi-definite covariance matrix $\sum_j$, an opacity $\alpha_j$, and view-dependent color coefficients $\mathbf{c}_j$.
Once the parameters of the 3D Gaussian splats $\mathcal{G}$ are determined, novel view synthesis can be achieved through alpha-blending: $\hat{\mathbf{I}}^p = \text{Render}(\mathcal{G},\mathbf{p})$.
We can use $\mathcal{I}$ to supervise the optimization of $\mathcal{G}$
\begin{align}
    \label{eq:3gds_loss}
    \arg \min_{\mathcal{G}} \lambda_1 \sum_{i=1}^N   \|(\mathbf{I}_i-\hat{\mathbf{I}}^{p_i})\|_1 + \lambda_2 \sum_{i=1}^N \text{D-SSIM}(\mathbf{I}_i,\hat{\mathbf{I}}^{p_i}), 
\end{align}
where $\mathbf{p}_i$ denotes the camera perspective of the image $\mathbf{I}_i$, $\hat{\mathbf{I}}^{p_i} = \text{Render}(\mathcal{G},\mathbf{p}_i)$, and $\lambda_1+\lambda_2=1$.
For novel view synthesis, given a camera perspective $\mathbf{p}$, the process involves the following steps: projecting each 3D Gaussian onto a 2D image plane, sorting the Gaussians by depth along the view direction, and blending the Gaussians from front to back for each pixel.
A key advantage of 3DGS \citep{kerbl20233d} is its ability to synthesize a new view in a single pass, whereas NeRF requires pixel-by-pixel rendering. This efficiency makes 3DGS particularly well-suited for time-sensitive 3D representation applications, offering a significant performance boost over NeRF.

\subsection{3D Gaussian Inpainting}
\label{subsec:3dgi}

Given a set of captured images $\mathcal{I}= \{\mathbf{I}_i\}_{i=1}^N$ and corresponding binary mask maps $\mathcal{M}= \{\mathbf{M}_i\}_{i=1}^N$ delineating objects for removal (See \Cref{fig:fig1}), 3D Gaussian Inpainting (3DGI) constructs a new 3D Gaussian splatting (3DGS) representation. This representation eliminates specified objects and replaces them with content that integrates with the environment.
The resulting 3DGS representation can synthesize arbitrary views where the specified objects are imperceptibly absent, maintaining visual coherence across viewpoints while effectively 'erasing' targeted objects.
We can use the segment anything model (SAM) \citep{kirillov2023segany} with few manual annotations to generate mask maps, aligning with methods like \citep{ye2023gaussian} for precise object delineation.

\textbf{SOTA methods and limitations.}
An intuitive approach to 3D Gaussian Inpainting (3DGI) involves deriving a 3D mask for the specified objects based on the provided 2D masks. The process of new view synthesis then follows a two-step procedure: first, generating the specified view and its corresponding mask, and then applying existing 2D image inpainting techniques to achieve the desired 3DGI effect. This methodology has been adopted in recent works by \citet{wang2024gaussianeditor} and \citet{ye2023gaussian}.
However, this approach does not leverage the complementary information available across multiple viewpoints during the inpainting process. A key example is the failure to utilize information from regions that may be occluded in one view but visible in another. Consequently, this method struggles to maintain consistency with the surrounding environment, particularly when dealing with large masked regions. This limitation underscores the need for more sophisticated techniques to effectively integrate and synthesize information from multiple perspectives to achieve more coherent and realistic 3D inpainting results.
Beyond this solution, the latest work \cite{liu2024infusion} utilizes the cross-view complementary cues through depth perception. It formulates the 3D Gaussian inpainting as two tasks, \ie, 2D image inpainting and depth inpainting, and the complementary cues in multiple views are implicitly utilized via depth projection. However, depth maps cannot fully represent complementary cues, such as the texture pattern from adjacent perspectives, and the depth project can hardly get high-quality depth maps when moving objects across different views. As case 2 shown in \Cref{fig:fig1}, InFusion synthesizes new views with obvious artifacts.

%%--------------------------------------------------------------------------------------------
% \section{VISTA: Visibility-Uncertainty-guided 3DGI via Scene Conceptual Learning}
\section{Methodology}
\label{sec:method}

This section details the proposed framework called VISibility-uncerTainty-guided 3D Gaussian inpainting via scene concepTional learning (VISTA). The core principle is to identify the visibility of 3D points across different views and utilize this information to guide the use of complementary visual and semantic cues for 3D Gaussian inpainting.

To elucidate this concept, we introduce the visibility-uncertainty-guided 3D Gaussian inpainting (VISTA-GI) in \Cref{subsec:vista-gi}, where we define the visibility uncertainty of 3D points and employ it to guide the use of complementary visual cues for 3DGI.
In \Cref{subsec:vista-cl}, we propose leveraging the visibility uncertainty to learn the semantic concept of the scene without specified objects. We then perform concept-driven Diffusion inpainting to process the input images, harnessing complementary semantic cues.
To fully utilize complementary visual and semantic cues, we propose in \Cref{subsec:vista} an iterative combination of VISTA-GI and VISTA-CL.
Finally, in \Cref{subsec:distrator}, we extend our VISTA framework to address the challenge of dynamic distractors in captured images. This extension excludes transient objects, resulting in clearer and more consistent novel view synthesis.

\subsection{VISTA-GI: Visibility-uncertainty-guided 3D Gaussian Inpainting}
\label{subsec:vista-gi}

\textbf{Initial 3D Gaussian Splatting.} Given the input images $\mathcal{I}=\{\mathbf{I}_i\}_{i=1}^N$, we employ the original 3DGS method in \Cref{subsec:3dgs} and \Cref{eq:3gds_loss} to construct a 3D representation $\mathcal{G}$. This representation can then be utilized to render novel views. However, as illustrated in \Cref{fig:pipeline}, this initial representation fails to exclude dynamic objects (such as fish) and exhibits noticeable artifacts, including blurring.

\textbf{Visibility uncertainty of 3D Points.} We define a set of adjacent camera perspectives/views denoted as $\mathcal{P}=\{\mathbf{p}_v\}_{v=1}^V$, where $V$ is the number of adjacent views. For a 3D point $\mathbf{X}$ in the scene, we can project it to different camera perspectives in $\mathcal{P}$ via the built 3DGS $\mathcal{G}$ and get their colors under $V$ views, \ie, $\{\mathbf{x}_v\}_{v=1}^V$.
Then, we calculate the variations of colors of the point under different views
\begin{align}
    \label{eq:vu}
    u_\mathbf{x} = \text{var}(\{\mathbf{x}_v\}_{v=1}^V),
\end{align}
where $\text{var}(\cdot)$ is the variation function. We denote the result $u_\mathbf{x}$ as the \textit{visibility uncertainty} of the 3D point $\mathbf{X}$. Intuitively, $u_\mathbf{x}$ represents the visibility and consistency of the point across the $V$ views. For example, if the point $\mathbf{X}$ can be seen at all views, the colors under different views are consistent and $u_\mathbf{x}$ is small. If the point can be only seen by a few views or its color deviates between different views, the visibility uncertainty tends to be significantly high.

\textbf{Reoptimized 3D Gaussian inpainting.} With the 3D point's visibility uncertainty, we aim to calculate the visibility uncertainty map of the input image and measure the visibility of each pixel at other views. Specifically, for an image $\mathbf{I}_i$ in $\mathcal{I}$, we first calculate its depth map $\mathbf{D}_i$ based on the $\mathcal{G}$. Then, we project each pixel of $\mathbf{I}_i$ to a 3D point and calculate its visibility uncertainty via \Cref{eq:vu} under $V$ adjacent views. Then, we obtain a pixel-wise visibility uncertainty map, which is normalized by dividing each pixel's uncertainty value by the standard deviation computed across all uncertainty values. The resulting normalized map is denoted as $\mathbf{U}_i$.
For the $N$ input images, we have $N$ visibility uncertainty maps $\mathcal{U} = \{\mathbf{U}_i\}_{i=1}^N$. 
Then, we use them to update the original mask maps $\mathcal{M}$ and uncertainty maps $\mathcal{U}$ by
\begin{align}
    \label{eq:updated_mask}
    \mathbf{M}_i' = \mathbf{U}_i \odot (1-\mathbf{M}_i) + \vartheta \cdot \mathbf{M}_i,
\end{align}
where the first term weights the unmasked regions via the visibility uncertainty map: the points other views cannot see should be assigned low weights during optimization. The $\vartheta$ controls the constraint degrees of the original masks.
Then, we obtain the finer mask maps $\{\mathbf{M}_i'\}_{i=1}^N$ and re-optimize the 3D representation by adding the guidance of mask maps to the objective function in \Cref{eq:3gds_loss}:
\begin{align}
    \label{eq:init_3gds_loss}
    & \arg \min_{\mathcal{G}} \left ( \lambda_1 \sum_{i=1}^N \|(1-\mathbf{M'}_i) \odot (\mathbf{I}_i-\hat{\mathbf{I}}^{p_i})\|_1 + \notag \right. \\ 
    &\left. \lambda_2 \sum_{i=1}^N \text{D-SSIM}(\mathbf{I}_i,\hat{\mathbf{I}}^{p_i},1-\mathbf{M'}_i) \right ),  
\end{align}
where we have $\hat{\mathbf{I}}^{p_i} = \text{Render}(\mathcal{G},\mathbf{p}_i) $ and $\lambda_1+\lambda_2=1$. Intuitively, the objective function is to ignore the mask and high-uncertainty regions during the optimization. As a result, we get an updated counterpart $\tilde{\mathcal{G}}$. Similar strategies have been also adopted in recent works \citep{sabourgoli2024spotlesssplats,sabour2023robustnerf}.

Intuitively, with the visibility uncertainty maps, we can exclude the pixels that other views cannot see to build the 3D representation, which explicitly leverages the complementary visual cues.
As the $\mathcal{U}$ shown in \Cref{fig:pipeline} (Bottom) , the pixels with high uncertainty denote the corresponding points (e.g., dynamic fishes) visible from only a few views. This is reasonable since the dynamic fishes are at different locations across different views. We also display the updated 3D representation $\tilde{\mathcal{G}}^1$, showing that the dynamic objects and some artifacts are removed. 

\subsection{VISTA-CL: Visibility-uncertainty-guided Scene Conceptual Learning}
\label{subsec:vista-cl}

VISTA-GI can reconstruct masked objects when complementary visual information is available from alternative viewpoints. However, for masked regions lacking such cues, we need a more sophisticated approach to comprehend the scene holistically and generate plausible new content to fill these gaps.
To achieve this, we propose to learn a conceptual representation $\mathbf{s}$ of the scene through textual inversion \citep{gal2022image,zhu2024cosalpure}, which can be formulated as 
\begin{align}   
    \label{eq:conceptlearn}
    \mathbf{s} = \text{ConceptLearn}(\mathcal{I},\mathcal{U},\mathcal{M}),
\end{align}
The learned concept $\mathbf{s}$ is a token and encapsulates the scene's essence without the masked objects. We then leverage $\mathbf{s}$ to process the input images, eliminating the masked objects
\begin{align}
    \label{eq:conceptinpaint}
    \tilde{\mathbf{I}}_i = \text{ConceptInpaint}(\mathbf{s},\mathbf{I}_i,\mathcal{U},\mathcal{M}), \forall \mathbf{I}_i \in \mathcal{I},
\end{align}

\textbf{Scene conceptual learning.} We formulate the scene conceptual learning, \ie, as the personalization text-to-image problem \citep{ruiz2023dreambooth} based on textual inversion \citep{gal2022image}, and we add the guidance of the visibility uncertainty maps in \Cref{subsec:3dgi}. Specifically, we have a pre-trained text-2-image diffusion model containing an image autoencoder with $\phi$ and $\phi^{-1}$ as encoder and decoder, a text encoder $\varphi$, and a conditional diffusion model $\epsilon_\theta$ at latent space. Then, we learn the scene concept $\mathbf{s}$ by optimizing the following objective function
%
%--------------------------
\begin{align}
\label{eq:conceptlearn1}
\textbf{s} = & \argmin\limits _{\textbf{s}^*} \mathbb{E}_{\mathbf{I}_i\in \mathcal{I}, \textbf{z} = \phi 
    (\mathbf{I}), \mathbf{y},\epsilon\in \mathcal{N}(0,1),t} ( \|(1-\mathbf{M}_i') \odot \notag \\
&(\epsilon_\theta(\mathbf{z}_t,t, \Upsilon(\varphi(\mathbf{y}),\textbf{s}^*))-\epsilon)\|_2^2),
\end{align}
%--------------------------
%
where $\mathbf{y}$ is a fixed text (\ie, `a photo of $S^*$') and the function $\Upsilon(\Gamma(\mathbf{y}),\textbf{s}^*)$ is to replace the token of `$S^*$' within $\Gamma(\mathbf{y})$ with $\textbf{s}^*$. 
The tensor $\mathbf{M}_i'$ is calculated via \Cref{eq:updated_mask} based on the visibility uncertainty map and the given mask map.
Intuitively, we use the \Cref{eq:conceptlearn1} to force the learned concept to mainly contain the unmasked scene regions.
To validate the learned concept, we can feed `a photo of $S^*$' to the T2I diffusion model to generate images about the learned concept.
As shown in \Cref{fig:pipeline}, the images in the lower right are created directly by the T2I diffusion model and illustrate a concept similar to the original scene without any dynamic objects. 
% Consequently, the diffusion model tends to inpaint in unknown areas with existing objects from the original scene during the inpainting process.

\textbf{Scene conceptual-guided inpainting.} We use the learned concept $\textbf{s}$ to inpaint all input images through the pre-trained T2I diffusion model. Given one image $\mathbf{I}$ from $\mathcal{I}$, we can extract its latent code by $\mathbf{z}=\phi(\mathbf{I})$. Then, we perform the forward diffusion process by iteratively adding Gaussian noise to the $\mathbf{z}$ over $T$ timesteps, obtaining a sequence of noisy latent codes, \ie, $\mathbf{z}_0$, $\mathbf{z}_1$, \ldots, $\mathbf{z}_T$, where $\mathbf{z}_0=\mathbf{z}$. At the $t$th step, the latent is obtained by
%--------------------------
\begin{align}
    \label{eq:forward}
    q(\mathbf{z}_t|\mathbf{z}_0) = \sqrt{\overline{\alpha}_t}\textbf{z}_{0} + \sqrt{1-\overline{\alpha}_t}\epsilon_t,~ \epsilon_t \sim \mathcal{N}(0, \mathbb{I}),
\end{align}
%-------------------------
where $\overline{\alpha}_t = \prod_{\tau=1}^{t}(1-\beta_\tau)$. $\mathcal{N}(0, \mathbb{I})$ represents the standard Gaussian distribution.
As we set the time step as $T$, the complete forward process can be expressed as 
%--------------------------
$   \mathbf{z}_{T} \sim \text{q}(\mathbf{z}_{1:T} | \mathbf{z}_0) = \prod_{t=1}^{T}{\text{q}(\textbf{z}_{t} | \mathbf{z}_{t-1})}.
$

At the reverse denoising process, we follow the strategy of RePaint \citep{lugmayr2022repaint} but embed the guidance of visibility uncertainty maps and the learned concept $\mathbf{s}$. 
Intuitively, at the time step $t>1$ during denoising, we only denoise the masked regions conditioned on the scene concept $\mathbf{s}$ while maintaining the unmasked regions with the same content in \Cref{{eq:forward}}, that is, we have
\begin{align}
    \tilde{\mathbf{z}}_{t-1} = (1-\mathbf{m}') \odot \mathbf{z}_{t-1} + \mathbf{m}' \odot \hat{\mathbf{z}}_{t-1},
\end{align}
where $\mathbf{z}_{t-1} \sim q(\mathbf{z}_t|\mathbf{z}_0)$ and $\mathbf{m}'$ is the downsampled $\mathbf{M}'\in \{\mathbf{M}_i'\}_{i=1}^N$ calculated by \Cref{eq:updated_mask} and has the exact resolution as the latent code $\mathbf{z}_{t-1}$.
$\hat{\mathbf{z}}_{t-1}$ is denoised from the $\tilde{\mathbf{z}}_{t}$ with the guidance of the learned concept $\mathbf{s}$, that is,
%
%--------------------------
\begin{align}
\hat{\mathbf{z}}_{t-1} =& \frac{1}{\sqrt{\alpha }_{t}}(\tilde{\mathbf{z}}_{t} - \frac{\beta_t}{\sqrt{1-\bar{\alpha}_{t} }}\epsilon_\theta(\tilde{\mathbf{z}}_{t}, t, \mathbf{s})) + \notag \\
&\sigma_t \xi, \text{s.t.}, \xi \sim \mathcal{N}(0, \mathbb{I}),
% \frac{\sqrt{\bar{\alpha }_{t-1}}(1-\alpha_t)}{1-\bar{\alpha}_t}\tilde{\mathbf{z}}_0
%  + \frac{\sqrt{\alpha }_{t}(1-\bar{\alpha}_{t-1})}{1-\bar{\alpha}_t}\hat{\mathbf{z}}_t+\sigma_t \xi,
    \label{eq:pred_zt-1}
\end{align}
%--------------------------
%
If $t=1$, $\tilde{\mathbf{z}}_0 = (1-\mathbf{m}') \odot \mathbf{z} + \mathbf{m}' \cdot \hat{\mathbf{z}}_{0}$. Then, we can get the inpainted image via decoder $\tilde{\mathbf{I}}= \phi^{-1}(\tilde{\mathbf{z}}_0)$.
We can use the above $\text{ConceptInpaint}$ to process each image within $\mathcal{I}$ and get a new image set $\tilde{\mathcal{I}}$.

\subsection{VISTA: Combining VISTA-GI and VISTA-CL}
\label{subsec:vista}

Given the input images $\mathcal{I}$ and their corresponding mask maps $\mathcal{M}$, VISTA-GI generates visibility uncertainty maps $\mathcal{U}$ as the visual cues and refines the 3DGS representation $\tilde{\mathcal{G}}$. VISTA-CL takes $\mathcal{I}$, $\mathcal{U}$, and $\mathcal{M}$ as inputs and produces processed input images $\tilde{\mathcal{I}}$ as the semantic cues.
Intuitively, we can combine the raw images $\mathcal{I}$ and $\tilde{\mathcal{I}}$, feed them back into VISTA-GI, where $\tilde{\mathcal{I}}$ serve as better views. This allows for an iterative process between VISTA-GI and VISTA-CL.
We denote the $k$-th iteration's 3D representation from VISTA-GI as $\tilde{\mathcal{G}}^k$ and the processed images from VISTA-CL as $\tilde{\mathcal{I}}^k$.

In practice, three iterations are typically sufficient to achieve smooth convergence of the training metrics. The hyperparameter $\vartheta$ is initialized by 0 and increases by 0.1 with each iteration.  We show an example in \Cref{fig:pipeline}. The synthetic views $\tilde{\mathcal{G}}^1$, $\tilde{\mathcal{G}}^2$, and $\tilde{\mathcal{G}}^3$ gradually contain fewer distractors, and the results of the final iteration $\tilde{\mathcal{G}}^3$ demonstrate clean and clear views, which means better 3D inpainting under the guidance of the visual and semantic cues.

%%--------------------------------------------------------------------------------------------
\begin{figure}[t]
\vspace{-5pt}
\centering
\includegraphics[width=0.99\linewidth]{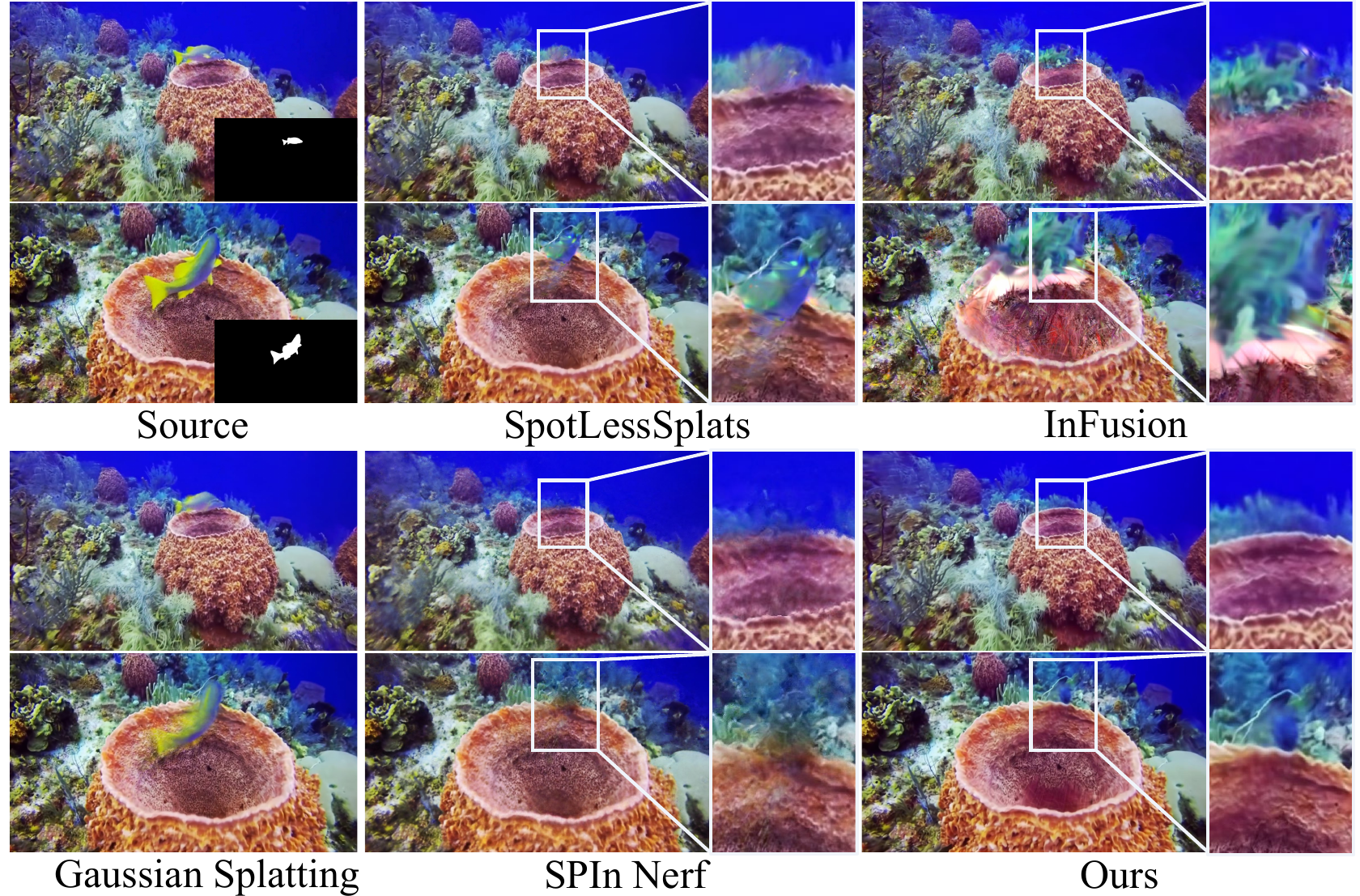}\\
% \caption{The figures illustrate the performance of various inpainting methods on our dataset, especially for dynamic objects. Some perspectives can compensate for some areas that need repair, while others require direct inpainting from the algorithm. This scene represents a scenario that can effectively reflect real-world inpainting task datasets.} 
\vspace{-5pt}
\caption{Example of dynamic inpainting on the Underwater 3D Inpainting Dataset.}
\label{fig:qualitative-Underwater}
\vspace{-10pt}
\end{figure}

\subsection{VISTA for Dynamic Distractor Removal}
\label{subsec:distrator}

VISTA could be easily extended to remove dynamic distractors across multi-view images $\mathcal{I}$ by identifying the dynamic regions in $\mathcal{I}$ and obtaining the mask maps $\mathcal{M}$. 
In our implementation, we use the tracking method and MASA \citep{li2024matching} to automatically get the mask maps for dynamic objects in the scene. MASA is an open-vocabulary video detection and segmentation model introducing coarse pixel-level information to our method. This plays a similar role as DEVA \citep{cheng2023tracking} used in Gaussian Grouping \citep{ye2023gaussian}. However, the masks used in Gaussian Grouping are limited to static objects, while we mask static and dynamic objects that need to be inpainted. 
For dynamic objects, the uncertainty map can complement the coarse mask that excludes those dynamic distractors from the reconstruction. 
As shown in \Cref{fig:pipeline},  the synthetic view  $\mathcal{G}$  obtained without masks fairly removes those fish moving greatly but ignores those objects without significant movement. The semantic information in the coarse masks $\mathcal{M}$ identifies these distractors, which the uncertainty map $\mathcal{U}$ cannot detect, and then these distractors can be eliminated by VISTA-CL. As a result, VISTA can remove both static and dynamic distractors in the scene by combining these two mask maps in \Cref{eq:updated_mask}. 

% using fig2
% As shown in \Cref{fig:mask_ablation}, the synthetic view obtained without masks fairly removes those fish moving greatly but ignores those objects without significant movement. The semantic information in the coarse masks identifies these distractors, which the uncertainty map cannot detect, and then these distractors can be eliminated by VISTA-CL. As a result, VISTA can remove both static and dynamic distractors in the scene by combining these two mask maps in \Cref{eq:updated_mask}. 

\section{Experiments}

\subsection{Datasets and Metrics}
To evaluate our method, we conduct experiments on the SPIn-NeRF Dataset and the Underwater 3D Inpainting Dataset for scene repair in different scenarios. 

\begin{table}[]
    \centering
\resizebox{0.7\linewidth}{!}{
    \begin{tabular}{l|ccc}
        \toprule
         Method&  UCIQE $\uparrow$ &URanker $\uparrow$ &CLIP Score $\uparrow$\\
         \midrule
         SPIn-NeRF  &  0.49&1.59&0.70\\
         InFusion   &0.50&1.52&0.71\\
         SpotLess   &0.50&1.59&0.70\\
         \midrule
         Ours&\textbf{0.51}&\textbf{1.64}&\textbf{0.72}\\
         \bottomrule
    \end{tabular}
}
\vspace{-0.5em}
\caption{Results of dynamic inpainting on the Underwater 3D Inpainting Dataset.\label{tab:Quantitative-Underwater}}
\vspace{-10px}
\end{table}

\textbf{Underwater 3D inpainting dataset.} This dataset is derived from the underwater object tracking dataset UTB180 \citep{alawode2022utb180}, from which we selected multiple videos for resampling, ultimately forming 10 underwater scene datasets. We resample the video in certain FPS to fulfill the motion requirements of initial reconstruction. Each scene contains dozens of images from various viewpoints, and the initial Structure from Motion point cloud and camera intrinsics are obtained via COLMAP \citep{schonberger2016structure}. Each viewpoint image undergoes object detection using the open-source method MASA \citep{li2024matching} to obtain rough object masks.
 
\textbf{SPIn-NeRF dataset.} The SPIn-NeRF dataset was proposed in \cite{mirzaei2023spin}. It contains 10 general 3D inpainting scenes, divided into 3 indoor and 7 outdoor scenes. Each scene includes 100 images from various viewpoints, along with corresponding masks. In these datasets, the ratio of the training set to the testing set is 6 to 4. We compare our method with other approaches using the provided camera intrinsics and initialized SfM point cloud.

\begin{table}[]
\vspace{0pt}
    \centering
    \resizebox{0.9\linewidth}{!}{
     \begin{tabular}{l|cccc}
        \toprule
        Method & LPIPS $\downarrow$& Fid $\downarrow$& PSNR $\uparrow$&SSIM $\uparrow$\\
        \midrule
        Masked Gaussians& 0.594&278.32 & 10.77&0.29\\
        SPIn-NeRF& 0.465& 156.64 & 15.80&0.46\\
        Gaussians Grouping& 0.454&123.48 & 14.86&0.27\\
 GScream& 0.422& 114.41& 15.98&\textbf{0.59}\\
        InFusion & 0.567&118.26& 15.59&0.53\\
        \midrule
        Ours& \textbf{0.418}&\textbf{113.58}&\textbf{16.48}&\textbf{0.59}\\
        \bottomrule
    \end{tabular}
    }
    \caption{Results of static inpainting on the SPIn-NeRF Dataset.}
    \label{tab:Quantitative-SPIn-NeRF}
\vspace{-10pt}
\end{table}

\textbf{Metrics.} Following SPIn-NeRF, we evaluate results in two quantitative terms: one for static scenes with ground truth using PSNR, SSIM, LPIPS, and Fid for Reference-based IQA (Image Quality Assessment), and the other for dynamic scenes without ground truth using UCIQE \citep{yang2015underwater}, URanker \citep{guo2023underwater} and CLIP Score \citep{hessel2021clipscore} for the underwater Non-Reference IQA. Following the typical comparison methods in SPIn-NeRF \citep{mirzaei2023spin} and RefFusion \citep{mirzaei2024reffusion}, LPIPS, and Fid are calculated around the masked region by considering the bounding box of the mask. UCIOE is a generally used underwater metric that utilizes a linear combination of chroma, saturation, and contrast for quantitative assessment, quantifying uneven color casts, blurriness, and low contrast. URanker is a transformer-based metric to assess the quality of underwater images. Meanwhile, the CLIP Score measures the relation between image and text. We serve `An underwater scene without fish' as the caption to evaluate the effects of fish removal.
% In addition, due to the limitation of the underwater Non-Reference IQA, we also show the qualitative visualization comparison of the various works by demonstrating their synthetic reconstructions.

\begin{figure}[t]
% \vspace{-5pt}
\centering
\includegraphics[width=1.0\linewidth]{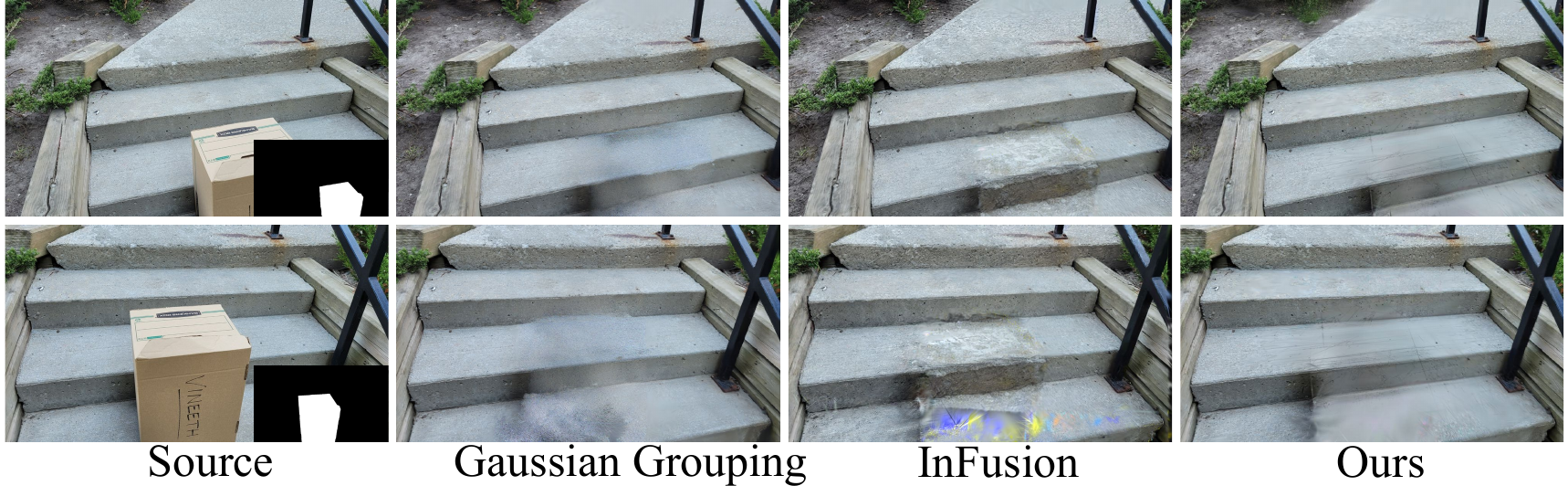}\\
\vspace{-0.5em}
\caption{Visualized examples of static inpainting on SPIn-NeRF.} 
\label{fig:qualitative-SPIn-NeRF}
\vspace{-1.5em}
\end{figure}

% By Yu: do not use unmeaningful names like "fig3.2.pdf" as name in the future. We need to use some meaningful names like "SPIn-NeRF-Viz.pdf" as name. 

\subsection{Experimental Results}

We compare our method with several state-of-the-art open-source 3D inpainting methods, such as Infusion \citep{liu2024infusion}, SPIn-NeRF \citep{mirzaei2023spin}, Gaussian Grouping \citep{ye2023gaussian}, and SpotLessSplats \citep{sabourgoli2024spotlesssplats}. SpotLessSplats is only designed for scenarios with dynamic distractors, while others are the latest static inpainting methods. Infusion \citep{liu2024infusion} and GScream \citep{wang2025learning} are retrained using their publicized code. 

\textbf{Results on underwater 3D inpainting dataset.} 
Our method demonstrates superior underwater inpainting performance in \Cref{fig:qualitative-Underwater} , achieving artifact-free 3D consistency across dynamic objects where baselines fail: SpotLessSplats incompletely removes moving fish Gaussians; Infusion induces viewpoint distortions despite localized clarity; SPIn-NeRF maintains 3D coherence but exhibits artifacts.
Quantitative validation in \Cref{tab:Quantitative-Underwater} reveals our approach surpasses competitors in UCIQE and URanker metrics through uncertainty-guided reconstruction, while superior CLIP Scores confirm effective target object removal through optimized semantic-geometric integration.

\textbf{Results on SPIn-NeRF dataset.} \Cref{fig:qualitative-SPIn-NeRF} depicts an example scene from the SPIn-NeRF Dataset masking a stationary box that requires inpainting. The results of Gaussian Grouping are fairly realistic at the 2D image level, but there are significant inconsistencies between perspectives, such as distortion at the edges of stairs. The results of InFusion appear more realistic from one certain perspective. Still, its approach of optimizing one single view compromises the performance of other perspectives, leading to unpredictable artifacts in those views. Our method benefits from an iterative progressive optimization approach, ensuring consistency across perspectives through multiple inpainting and reconstruction, resulting in more stable outcomes.

\begin{table}[]
\vspace{-10pt}
    \centering
    \resizebox{0.8\linewidth}{!}{
    \begin{tabular}{l|ccc}
        \toprule
         Method&  UCIQE $\uparrow$ &URanker $\uparrow$ &CLIP Score $\uparrow$\\
         \midrule
         Ours w/o VISTA-GI&  0.48&1.52&0.70\\
         Ours w/o VISTA-CL&0.50&1.59&0.69\\
         \midrule
         Ours&\textbf{0.51}&\textbf{1.64}&\textbf{0.72}\\
         \bottomrule
    \end{tabular}
    }
    \caption{Ablation study of VISTA-GI and VISTA-CL on the Underwater 3D Inpainting Dataset.}
    \label{tab:vista_ablation}
\vspace{0em}
\end{table}

\textbf{Ablation study on VISTA-GI and VISTA-CL.} 
We conducted the ablation study on the underwater 3D inpainting dataset by removing the VISTA-GI and VISTA-CL from our pipeline respectively. The specific results are shown in \Cref{tab:vista_ablation}. 
Our experiments demonstrate two key findings. First, attempting reconstruction using only a 2D generative model without VISTA-GI leads to significantly degraded metrics. This validates that VISTA-GI's uncertainty guidance effectively mitigates multi-view inconsistencies during 3D reconstruction, resulting in higher-quality outputs.
Second, while omitting VISTA-CL maintains image quality comparable to existing methods like SplotLess and SPIn-NeRF, the lack of concept-guided learning significantly reduces CLIP-Score metrics. This indicates that without conceptual constraints, the inpainting process produces results that are visually plausible but semantically inconsistent with the scene context.

\begin{figure}[t]
\vspace{0pt}
\centering
\includegraphics[width=\linewidth]{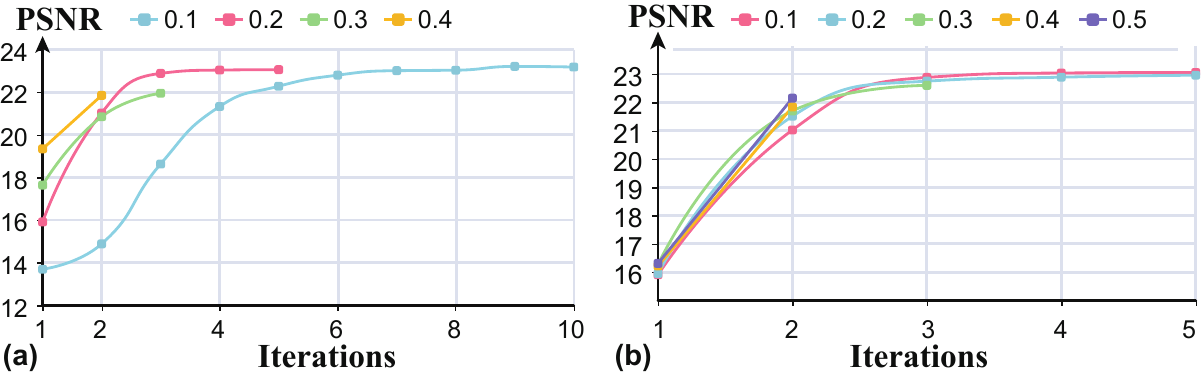}
\vspace{-15pt}
\caption{(a) Relationship between 3DGS rendering quality and noise reduction ratios in diffusion. (b) Relationship between 3DGS rendering quality and increasing ratio of $\vartheta$.}
\label{fig:decreasingratio}
\vspace{-20pt}
\end{figure}

% During diffusion model inference, we investigate how different noise reduction strategies affect reconstruction quality. Starting from an initial noise strength of 1.0, we systematically decrease the noise at each iteration by a fixed ratio. We evaluate four different reduction ratios $\{0.1, 0.2, 0.3, 0.4\}$ and analyze their impact on reconstruction quality across iterations using our dataset.
% As shown in \Cref{fig:decreasingratio} (a), while all ratios lead to improved PSNR values over iterations, the reduction ratio of 0.2 achieves optimal convergence in the fewest iterations. Based on this empirical analysis, we adopt 0.2 as the noise reduction ratio in our method.
\textbf{Impact of noise reduction ratios in diffusion.} 
We analyze noise reduction strategies during diffusion model inference by implementing fixed-ratio iterative noise reduction from 1.0, evaluating four ratios $\{0.1, 0.2, 0.3, 0.4\}$ on reconstruction quality. As shown in \Cref{fig:decreasingratio} (a), all ratios improve PSNR iteratively, but the 0.2 ratio achieves optimal convergence fastest. Empirical validation confirms 0.2 as our method's optimal reduction ratio. 

% We use $\vartheta$ to control the prior constraint of the original masks.  We investigate how different $\vartheta$ increasing strategies affect reconstruction quality. The hyperparameter $\vartheta$ is initialized by 0 and increases by 0.1 with each iteration in our paper. We evaluate five different increase ratios $\{0.1, 0.2, 0.3, 0.4, 0.5\}$ and analyze their impact on reconstruction quality across iterations using our dataset. 
% As shown in \Cref{fig:decreasingratio} (b), all ratios lead to improved PSNR values over iterations. In the first two iterations, a higher increase ratio improves the reconstruction performance. However, an increase ratio above 0.1 indicates that the algorithm becomes overly confident in the inpainting areas too early, resulting in insufficient interaction of geometric and semantic information between the VISTA-GI and VISTA-CL modules, which subsequently leads to a decline in reconstruction performance in later iterations.
\textbf{Impacts of $\vartheta$ in \reqref{eq:updated_mask}.} 
We analyze mask prior constraint control via $\vartheta$ increase strategies (0.1-0.5), initialized at $\vartheta=0$ with 0.1 per-iteration increments. \Cref{fig:decreasingratio} (b) demonstrates iterative PSNR improvements across all ratios. While higher ratios enhance early-stage reconstruction, values bigger than 0.1 trigger premature inpainting confidence, causing insufficient VISTA-GI/CL module interaction and subsequent performance degradation. 

\textbf{Ablation study of large viewpoint differences.}
To evaluate the impact of viewpoint difference, we capture 34 images from continuously distributed viewpoints around a scene to create a ground truth (GT) 3DGS model. We systematically reduce the number of viewpoints by sampling them at different intervals $\{2,3,4,5,6,7\}$, where larger intervals represent larger viewpoint differences. 
We reconstruct a 3DGS model per sampling interval and establish a quantitative analysis framework linking viewpoint differences to reconstruction quality by benchmarking rendered images against GT models using standard metrics. Our method ensures robust reconstruction under significant viewpoint variations through view-consistency detection and repair mechanisms, while metrics degrade in extreme scenarios with missing key viewpoints due to irreversible information loss. 
% For each sampling interval, we construct a new 3DGS model and assess its quality by comparing its rendered images against those from the GT model using standard metrics. This methodology allows us to analyze how viewpoint difference affects reconstruction quality quantitatively.
% Our method maintains robust 3DGS reconstruction quality despite significant viewpoint variations by detecting and repairing inter-view inconsistencies, though extreme cases lacking key viewpoints experience metric degradation due to irrecoverable complementary information loss.

\begin{table}[hbt]
\vspace{-8pt}
    \centering
    \resizebox{0.7\linewidth}{!}{
     \begin{tabular}{c|ccc}
        \toprule
        Sampling interval& LPIPS $\downarrow$& PSNR $\uparrow$&SSIM $\uparrow$\\
        \midrule
        2& 0.09& 26.25&0.89\\
        3& 0.14& 23.42&0.83\\
        4& 0.16& 22.42&0.80\\
        5& 0.27& 18.09&0.65\\
        6& 0.25& 18.71&0.69\\
        7& 0.41& 15.66&0.57\\
        \bottomrule
    \end{tabular}
    }
    \vspace{-8pt}
    \caption{Quantitative results of large viewpoint differences. \label{tab:viewpoint_ablation}}
\vspace{-13pt}
\end{table}

% Considering that the reduction in available viewpoints for the training leads to decreased 3DGS reconstruction quality, our method still achieves good results even with significant viewpoint variation. This validates that our approach can detect inconsistencies between viewpoints and repair those areas despite the large viewpoint differences. However, in extreme cases, the absence of key viewpoints results in a loss of critical complementary information between viewpoints, leading to a significant decline in the scene's reconstruction metrics.

\textbf{Comparisons of different methods in extreme cases.}
To validate our advantages in the extreme case with large viewpoint differences, we quantitatively evaluate various methods for the extreme case, and the results are as \Cref{tab:extreme_case}. Our method still outperforms existing methods in removing dynamic distractors under such extreme conditions.

\begin{table}[hbt]
\vspace{-10pt}
    \centering 
    \resizebox{0.6\linewidth}{!}{
     \begin{tabular}{l|ccc}
        \toprule
        Method & LPIPS $\downarrow$& PSNR $\uparrow$&SSIM $\uparrow$\\
        \midrule
        InFusion& 0.23& 19.34&0.78\\
        SPIn-NeRF& 0.15& 23.33&0.82\\
        SpotLess& 0.14& 24.75&0.84\\
        \midrule
        Ours& \textbf{0.10}&\textbf{26.38}&\textbf{0.86}\\
        \bottomrule
    \end{tabular}
    }
    \vspace{-8pt}
    \caption{Comparison of different methods in extreme case. \label{tab:extreme_case}}
    \vspace{-10pt}
\end{table}

\begin{figure}
\vspace{-10pt}
    \centering
    \includegraphics[width=0.75\linewidth]{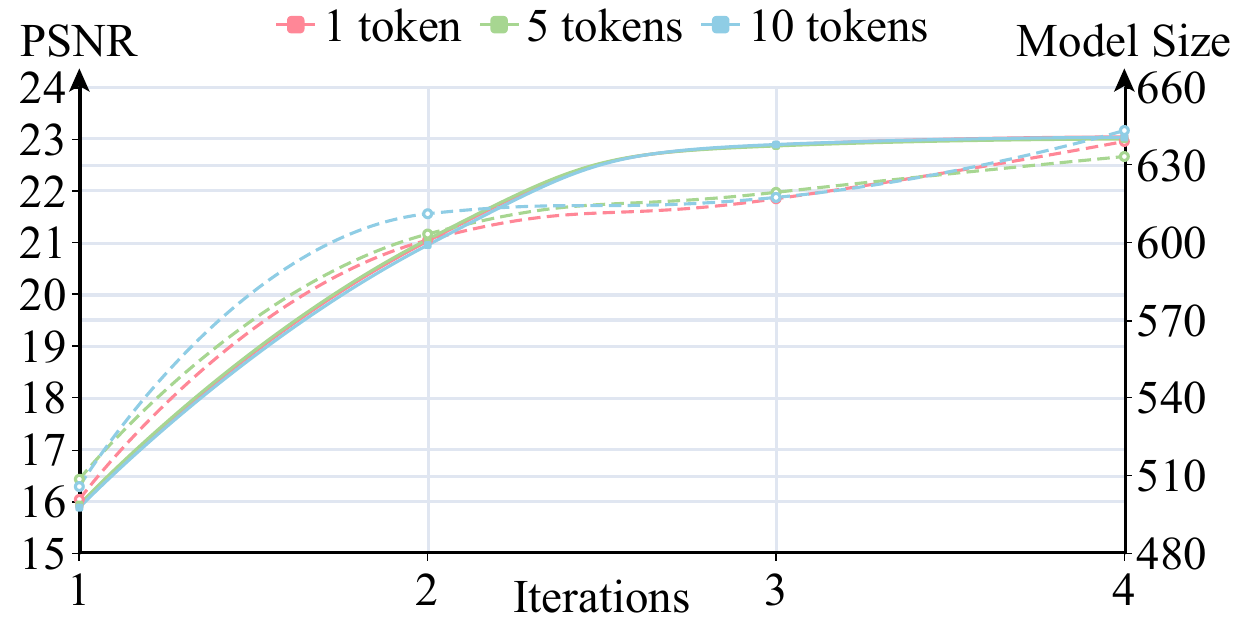}
    \vspace{-1em}
    \caption{Relationship between model performance (PSNR) and model size (MB) with different token numbers. The dashed and solid lines represent the model size and performance variations respectively. The model performance (solid lines) under different token numbers almost overlaps.}
    \label{iteration_charts}
\vspace{-18pt}
\end{figure}

\textbf{Effects of token numbers to depict one scene.} 
As shown in \Cref{iteration_charts}, our method balances scene depiction efficiency by optimizing token quantity (linked to color specificity) and textual inversion for coarse semantic learning, as excessive tokens inflate model size without performance gains. Training stabilizes at three iterations (determined by PSNR smoothing), beyond which redundant Gaussians overfit diffusion noise, further increasing model parameters.
% The quantity of tokens needed to describe a scene is important since each count corresponds to a specific color. The \Cref{iteration_charts} shows how the inpainting results change as the descriptive tokens change. Too many tokens to depict the scene do not increase the model performance and may even increase the model size. The effect of textual inversion is to focus on learning the rough semantic features of the scene rather than the detailed object features, thereby not necessarily requiring very detailed tokens.
% We also observed that the training PSNR became smooth after three iterations, which inspired us to set three iterations. More iterations cause a larger model size, which means excessive Gaussians to fit the noise introduced by the diffusion model. 

% The IQA results by our method are overall the best, as shown in \Cref{tab:Quantitative-SPIn-NeRF}.  

%%--------------------------------------------------------------------------------------------
\section{Conclusion}

In this work, we presented VISTA, a novel framework for 3D Gaussian inpainting that effectively leverages complementary visual and semantic cues from multiple input views. By introducing visibility uncertainty maps and combining visibility-uncertainty-guided 3D Gaussian inpainting (VISTA-GI) with scene conceptual learning (VISTA-CL), our method addresses key challenges in 3D scene editing for static and dynamic scenes. Experimental results on the SPIn-NeRF and UTB180-derived datasets demonstrate VISTA's superior performance over state-of-the-art techniques in generating high-quality 3D representations with seamlessly filled masked regions and effectively removing distractors. The versatility of our approach extends to handling complex inpainting scenarios and dynamic distractor removal, making it a powerful tool for various applications in augmented and virtual reality. By simultaneously leveraging geometric and conceptual information, VISTA represents a significant advancement in 3D Gaussian inpainting, bringing us closer to achieving seamless and realistic 3D scene editing and paving the way for more immersive virtual experiences.

{
    \small
    \bibliographystyle{ieeenat_fullname}
    \bibliography{main}

\begin{thebibliography}{43}
\providecommand{\natexlab}[1]{#1}
\providecommand{\url}[1]{\texttt{#1}}
\expandafter\ifx\csname urlstyle\endcsname\relax
  \providecommand{\doi}[1]{doi: #1}\else
  \providecommand{\doi}{doi: \begingroup \urlstyle{rm}\Url}\fi

\bibitem[Alawode et~al.(2022)Alawode, Guo, Ummar, Werghi, Dias, Mian, and Javed]{alawode2022utb180}
Basit Alawode, Yuhang Guo, Mehnaz Ummar, Naoufel Werghi, Jorge Dias, Ajmal Mian, and Sajid Javed.
\newblock Utb180: A high-quality benchmark for underwater tracking.
\newblock In \emph{Proceedings of the Asian Conference on Computer Vision}, pages 3326--3342, 2022.

\bibitem[Barron et~al.(2021)Barron, Mildenhall, Tancik, Hedman, Martin-Brualla, and Srinivasan]{barron2021mip}
Jonathan~T Barron, Ben Mildenhall, Matthew Tancik, Peter Hedman, Ricardo Martin-Brualla, and Pratul~P Srinivasan.
\newblock Mip-nerf: A multiscale representation for anti-aliasing neural radiance fields.
\newblock In \emph{Proceedings of the IEEE/CVF international conference on computer vision}, pages 5855--5864, 2021.

\bibitem[Barron et~al.(2022)Barron, Mildenhall, Verbin, Srinivasan, and Hedman]{barron2022mip}
Jonathan~T Barron, Ben Mildenhall, Dor Verbin, Pratul~P Srinivasan, and Peter Hedman.
\newblock Mip-nerf 360: Unbounded anti-aliased neural radiance fields.
\newblock In \emph{Proceedings of the IEEE/CVF conference on computer vision and pattern recognition}, pages 5470--5479, 2022.

\bibitem[Barron et~al.(2023)Barron, Mildenhall, Verbin, Srinivasan, and Hedman]{barron2023zip}
Jonathan~T Barron, Ben Mildenhall, Dor Verbin, Pratul~P Srinivasan, and Peter Hedman.
\newblock Zip-nerf: Anti-aliased grid-based neural radiance fields.
\newblock In \emph{Proceedings of the IEEE/CVF International Conference on Computer Vision}, pages 19697--19705, 2023.

\bibitem[Cheng et~al.(2023)Cheng, Oh, Price, Schwing, and Lee]{cheng2023tracking}
Ho~Kei Cheng, Seoung~Wug Oh, Brian Price, Alexander Schwing, and Joon-Young Lee.
\newblock Tracking anything with decoupled video segmentation.
\newblock In \emph{Proceedings of the IEEE/CVF International Conference on Computer Vision}, pages 1316--1326, 2023.

\bibitem[Gal et~al.(2022)Gal, Alaluf, Atzmon, Patashnik, Bermano, Chechik, and Cohen-Or]{gal2022image}
Rinon Gal, Yuval Alaluf, Yuval Atzmon, Or Patashnik, Amit~H Bermano, Gal Chechik, and Daniel Cohen-Or.
\newblock An image is worth one word: Personalizing text-to-image generation using textual inversion.
\newblock \emph{arXiv preprint arXiv:2208.01618}, 2022.

\bibitem[Goodfellow et~al.(2014)Goodfellow, Pouget-Abadie, Mirza, Xu, Warde-Farley, Ozair, Courville, and Bengio]{goodfellow2014generative}
Ian Goodfellow, Jean Pouget-Abadie, Mehdi Mirza, Bing Xu, David Warde-Farley, Sherjil Ozair, Aaron Courville, and Yoshua Bengio.
\newblock Generative adversarial nets.
\newblock \emph{Advances in neural information processing systems}, 27, 2014.

\bibitem[Guo et~al.(2023)Guo, Wu, Jin, Han, Zhang, Chai, and Li]{guo2023underwater}
Chunle Guo, Ruiqi Wu, Xin Jin, Linghao Han, Weidong Zhang, Zhi Chai, and Chongyi Li.
\newblock Underwater ranker: Learn which is better and how to be better.
\newblock In \emph{Proceedings of the AAAI conference on artificial intelligence}, pages 702--709, 2023.

\bibitem[Hessel et~al.(2021)Hessel, Holtzman, Forbes, Bras, and Choi]{hessel2021clipscore}
Jack Hessel, Ari Holtzman, Maxwell Forbes, Ronan~Le Bras, and Yejin Choi.
\newblock Clipscore: A reference-free evaluation metric for image captioning.
\newblock \emph{arXiv preprint arXiv:2104.08718}, 2021.

\bibitem[Ho et~al.(2020)Ho, Jain, and Abbeel]{ho2020denoising}
Jonathan Ho, Ajay Jain, and Pieter Abbeel.
\newblock Denoising diffusion probabilistic models.
\newblock \emph{Advances in neural information processing systems}, 33:\penalty0 6840--6851, 2020.

\bibitem[Kerbl et~al.(2023)Kerbl, Kopanas, Leimk{\"u}hler, and Drettakis]{kerbl20233d}
Bernhard Kerbl, Georgios Kopanas, Thomas Leimk{\"u}hler, and George Drettakis.
\newblock 3d gaussian splatting for real-time radiance field rendering.
\newblock \emph{ACM Trans. Graph.}, 42\penalty0 (4):\penalty0 139--1, 2023.

\bibitem[Kirillov et~al.(2023)Kirillov, Mintun, Ravi, Mao, Rolland, Gustafson, Xiao, Whitehead, Berg, Lo, Doll{\'a}r, and Girshick]{kirillov2023segany}
Alexander Kirillov, Eric Mintun, Nikhila Ravi, Hanzi Mao, Chloe Rolland, Laura Gustafson, Tete Xiao, Spencer Whitehead, Alexander~C. Berg, Wan-Yen Lo, Piotr Doll{\'a}r, and Ross Girshick.
\newblock Segment anything.
\newblock \emph{arXiv:2304.02643}, 2023.

\bibitem[Kutulakos and Seitz(2000)]{kutulakos2000theory}
Kiriakos~N Kutulakos and Steven~M Seitz.
\newblock A theory of shape by space carving.
\newblock \emph{International journal of computer vision}, 38:\penalty0 199--218, 2000.

\bibitem[Li et~al.(2024)Li, Ke, Danelljan, Piccinelli, Segu, Van~Gool, and Yu]{li2024matching}
Siyuan Li, Lei Ke, Martin Danelljan, Luigi Piccinelli, Mattia Segu, Luc Van~Gool, and Fisher Yu.
\newblock Matching anything by segmenting anything.
\newblock In \emph{Proceedings of the IEEE/CVF Conference on Computer Vision and Pattern Recognition}, pages 18963--18973, 2024.

\bibitem[Li et~al.(2022)Li, Lin, Zhou, Qi, Wang, and Jia]{li2022mat}
Wenbo Li, Zhe Lin, Kun Zhou, Lu Qi, Yi Wang, and Jiaya Jia.
\newblock Mat: Mask-aware transformer for large hole image inpainting.
\newblock In \emph{Proceedings of the IEEE/CVF conference on computer vision and pattern recognition}, pages 10758--10768, 2022.

\bibitem[Liu et~al.(2022)Liu, Shen, Chen, et~al.]{liu2022nerf}
Hao-Kang Liu, I Shen, Bing-Yu Chen, et~al.
\newblock Nerf-in: Free-form nerf inpainting with rgb-d priors.
\newblock \emph{arXiv preprint arXiv:2206.04901}, 2022.

\bibitem[Liu et~al.(2024)Liu, Ouyang, Wang, Cheng, Xiao, Zhu, Xue, Liu, Shen, and Cao]{liu2024infusion}
Zhiheng Liu, Hao Ouyang, Qiuyu Wang, Ka~Leong Cheng, Jie Xiao, Kai Zhu, Nan Xue, Yu Liu, Yujun Shen, and Yang Cao.
\newblock Infusion: Inpainting 3d gaussians via learning depth completion from diffusion prior.
\newblock \emph{arXiv preprint arXiv:2404.11613}, 2024.

\bibitem[Lombardi et~al.(2019)Lombardi, Simon, Saragih, Schwartz, Lehrmann, and Sheikh]{lombardi2019neural}
Stephen Lombardi, Tomas Simon, Jason Saragih, Gabriel Schwartz, Andreas Lehrmann, and Yaser Sheikh.
\newblock Neural volumes: Learning dynamic renderable volumes from images.
\newblock \emph{arXiv preprint arXiv:1906.07751}, 2019.

\bibitem[Lu et~al.(2024)Lu, Yu, Xu, Xiangli, Wang, Lin, and Dai]{lu2024scaffold}
Tao Lu, Mulin Yu, Linning Xu, Yuanbo Xiangli, Limin Wang, Dahua Lin, and Bo Dai.
\newblock Scaffold-gs: Structured 3d gaussians for view-adaptive rendering.
\newblock In \emph{Proceedings of the IEEE/CVF Conference on Computer Vision and Pattern Recognition}, pages 20654--20664, 2024.

\bibitem[Lugmayr et~al.(2022)Lugmayr, Danelljan, Romero, Yu, Timofte, and Van~Gool]{lugmayr2022repaint}
Andreas Lugmayr, Martin Danelljan, Andres Romero, Fisher Yu, Radu Timofte, and Luc Van~Gool.
\newblock Repaint: Inpainting using denoising diffusion probabilistic models.
\newblock In \emph{Proceedings of the IEEE/CVF conference on computer vision and pattern recognition}, pages 11461--11471, 2022.

\bibitem[Mildenhall et~al.(2021)Mildenhall, Srinivasan, Tancik, Barron, Ramamoorthi, and Ng]{mildenhall2021nerf}
Ben Mildenhall, Pratul~P Srinivasan, Matthew Tancik, Jonathan~T Barron, Ravi Ramamoorthi, and Ren Ng.
\newblock Nerf: Representing scenes as neural radiance fields for view synthesis.
\newblock \emph{Communications of the ACM}, 65\penalty0 (1):\penalty0 99--106, 2021.

\bibitem[Mirzaei et~al.(2023)Mirzaei, Aumentado-Armstrong, Derpanis, Kelly, Brubaker, Gilitschenski, and Levinshtein]{mirzaei2023spin}
Ashkan Mirzaei, Tristan Aumentado-Armstrong, Konstantinos~G Derpanis, Jonathan Kelly, Marcus~A Brubaker, Igor Gilitschenski, and Alex Levinshtein.
\newblock Spin-nerf: Multiview segmentation and perceptual inpainting with neural radiance fields.
\newblock In \emph{Proceedings of the IEEE/CVF Conference on Computer Vision and Pattern Recognition}, pages 20669--20679, 2023.

\bibitem[Mirzaei et~al.(2024)Mirzaei, De~Lutio, Kim, Acuna, Kelly, Fidler, Gilitschenski, and Gojcic]{mirzaei2024reffusion}
Ashkan Mirzaei, Riccardo De~Lutio, Seung~Wook Kim, David Acuna, Jonathan Kelly, Sanja Fidler, Igor Gilitschenski, and Zan Gojcic.
\newblock Reffusion: Reference adapted diffusion models for 3d scene inpainting.
\newblock \emph{arXiv preprint arXiv:2404.10765}, 2024.

\bibitem[Rombach et~al.(2022)Rombach, Blattmann, Lorenz, Esser, and Ommer]{rombach2022high}
Robin Rombach, Andreas Blattmann, Dominik Lorenz, Patrick Esser, and Bj{\"o}rn Ommer.
\newblock High-resolution image synthesis with latent diffusion models.
\newblock In \emph{Proceedings of the IEEE/CVF conference on computer vision and pattern recognition}, pages 10684--10695, 2022.

\bibitem[Ruiz et~al.(2023)Ruiz, Li, Jampani, Pritch, Rubinstein, and Aberman]{ruiz2023dreambooth}
Nataniel Ruiz, Yuanzhen Li, Varun Jampani, Yael Pritch, Michael Rubinstein, and Kfir Aberman.
\newblock Dreambooth: Fine tuning text-to-image diffusion models for subject-driven generation.
\newblock In \emph{Proceedings of the IEEE/CVF conference on computer vision and pattern recognition}, pages 22500--22510, 2023.

\bibitem[Ru{\v{z}}i{\'c} and Pi{\v{z}}urica(2014)]{ruvzic2014context}
Tijana Ru{\v{z}}i{\'c} and Aleksandra Pi{\v{z}}urica.
\newblock Context-aware patch-based image inpainting using markov random field modeling.
\newblock \emph{IEEE transactions on image processing}, 24\penalty0 (1):\penalty0 444--456, 2014.

\bibitem[Sabour et~al.(2023)Sabour, Vora, Duckworth, Krasin, Fleet, and Tagliasacchi]{sabour2023robustnerf}
Sara Sabour, Suhani Vora, Daniel Duckworth, Ivan Krasin, David~J Fleet, and Andrea Tagliasacchi.
\newblock Robustnerf: Ignoring distractors with robust losses.
\newblock In \emph{Proceedings of the IEEE/CVF Conference on Computer Vision and Pattern Recognition}, pages 20626--20636, 2023.

\bibitem[Sabour et~al.(2024)Sabour, Goli, Kopanas, Matthews, Lagun, Guibas, Jacobson, Fleet, and Tagliasacchi]{sabourgoli2024spotlesssplats}
Sara Sabour, Lily Goli, George Kopanas, Mark Matthews, Dmitry Lagun, Leonidas Guibas, Alec Jacobson, David~J. Fleet, and Andrea Tagliasacchi.
\newblock {SpotLessSplats}: Ignoring distractors in 3d gaussian splatting.
\newblock \emph{arXiv:2406.20055}, 2024.

\bibitem[Schonberger and Frahm(2016)]{schonberger2016structure}
Johannes~L Schonberger and Jan-Michael Frahm.
\newblock Structure-from-motion revisited.
\newblock In \emph{Proceedings of the IEEE conference on computer vision and pattern recognition}, pages 4104--4113, 2016.

\bibitem[Sohl-Dickstein et~al.(2015)Sohl-Dickstein, Weiss, Maheswaranathan, and Ganguli]{sohl2015deep}
Jascha Sohl-Dickstein, Eric Weiss, Niru Maheswaranathan, and Surya Ganguli.
\newblock Deep unsupervised learning using nonequilibrium thermodynamics.
\newblock In \emph{International conference on machine learning}, pages 2256--2265. PMLR, 2015.

\bibitem[Song et~al.(2020)Song, Sohl-Dickstein, Kingma, Kumar, Ermon, and Poole]{song2020score}
Yang Song, Jascha Sohl-Dickstein, Diederik~P Kingma, Abhishek Kumar, Stefano Ermon, and Ben Poole.
\newblock Score-based generative modeling through stochastic differential equations.
\newblock \emph{arXiv preprint arXiv:2011.13456}, 2020.

\bibitem[Suvorov et~al.(2022)Suvorov, Logacheva, Mashikhin, Remizova, Ashukha, Silvestrov, Kong, Goka, Park, and Lempitsky]{suvorov2022resolution}
Roman Suvorov, Elizaveta Logacheva, Anton Mashikhin, Anastasia Remizova, Arsenii Ashukha, Aleksei Silvestrov, Naejin Kong, Harshith Goka, Kiwoong Park, and Victor Lempitsky.
\newblock Resolution-robust large mask inpainting with fourier convolutions.
\newblock In \emph{Proceedings of the IEEE/CVF winter conference on applications of computer vision}, pages 2149--2159, 2022.

\bibitem[Tang et~al.(2023)Tang, Ren, Zhou, Liu, and Zeng]{tang2023dreamgaussian}
Jiaxiang Tang, Jiawei Ren, Hang Zhou, Ziwei Liu, and Gang Zeng.
\newblock Dreamgaussian: Generative gaussian splatting for efficient 3d content creation.
\newblock \emph{arXiv preprint arXiv:2309.16653}, 2023.

\bibitem[Tewari et~al.(2022)Tewari, Thies, Mildenhall, Srinivasan, Tretschk, Yifan, Lassner, Sitzmann, Martin-Brualla, Lombardi, et~al.]{tewari2022advances}
Ayush Tewari, Justus Thies, Ben Mildenhall, Pratul Srinivasan, Edgar Tretschk, Wang Yifan, Christoph Lassner, Vincent Sitzmann, Ricardo Martin-Brualla, Stephen Lombardi, et~al.
\newblock Advances in neural rendering.
\newblock In \emph{Computer Graphics Forum}, pages 703--735. Wiley Online Library, 2022.

\bibitem[Wang et~al.(2024)Wang, Fang, Zhang, Xie, and Tian]{wang2024gaussianeditor}
Junjie Wang, Jiemin Fang, Xiaopeng Zhang, Lingxi Xie, and Qi Tian.
\newblock Gaussianeditor: Editing 3d gaussians delicately with text instructions.
\newblock In \emph{Proceedings of the IEEE/CVF Conference on Computer Vision and Pattern Recognition}, pages 20902--20911, 2024.

\bibitem[Wang et~al.(2021)Wang, Liu, Liu, Theobalt, Komura, and Wang]{wang2021neus}
Peng Wang, Lingjie Liu, Yuan Liu, Christian Theobalt, Taku Komura, and Wenping Wang.
\newblock Neus: Learning neural implicit surfaces by volume rendering for multi-view reconstruction.
\newblock \emph{arXiv preprint arXiv:2106.10689}, 2021.

\bibitem[Wang et~al.(2025)Wang, Wu, Zhang, and Xu]{wang2025learning}
Yuxin Wang, Qianyi Wu, Guofeng Zhang, and Dan Xu.
\newblock Learning 3d geometry and feature consistent gaussian splatting for object removal.
\newblock In \emph{European Conference on Computer Vision}, pages 1--17. Springer, 2025.

\bibitem[Weder et~al.(2023)Weder, Garcia-Hernando, Monszpart, Pollefeys, Brostow, Firman, and Vicente]{weder2023removing}
Silvan Weder, Guillermo Garcia-Hernando, Aron Monszpart, Marc Pollefeys, Gabriel~J Brostow, Michael Firman, and Sara Vicente.
\newblock Removing objects from neural radiance fields.
\newblock In \emph{Proceedings of the IEEE/CVF Conference on Computer Vision and Pattern Recognition}, pages 16528--16538, 2023.

\bibitem[Wu et~al.(2024)Wu, Yi, Fang, Xie, Zhang, Wei, Liu, Tian, and Wang]{wu20244d}
Guanjun Wu, Taoran Yi, Jiemin Fang, Lingxi Xie, Xiaopeng Zhang, Wei Wei, Wenyu Liu, Qi Tian, and Xinggang Wang.
\newblock 4d gaussian splatting for real-time dynamic scene rendering.
\newblock In \emph{Proceedings of the IEEE/CVF Conference on Computer Vision and Pattern Recognition (CVPR)}, pages 20310--20320, 2024.

\bibitem[Yang and Sowmya(2015)]{yang2015underwater}
Miao Yang and Arcot Sowmya.
\newblock An underwater color image quality evaluation metric.
\newblock \emph{IEEE Transactions on Image Processing}, 24\penalty0 (12):\penalty0 6062--6071, 2015.

\bibitem[Ye et~al.(2024)Ye, Danelljan, Yu, and Ke]{ye2023gaussian}
Mingqiao Ye, Martin Danelljan, Fisher Yu, and Lei Ke.
\newblock Gaussian grouping: Segment and edit anything in 3d scenes.
\newblock In \emph{European Conference on Computer Vision}, 2024.

\bibitem[Yu et~al.(2018)Yu, Lin, Yang, Shen, Lu, and Huang]{yu2018generative}
Jiahui Yu, Zhe Lin, Jimei Yang, Xiaohui Shen, Xin Lu, and Thomas~S Huang.
\newblock Generative image inpainting with contextual attention.
\newblock In \emph{Proceedings of the IEEE conference on computer vision and pattern recognition}, pages 5505--5514, 2018.

\bibitem[Zhu et~al.(2024)Zhu, Guo, Juefei-Xu, Huang, Liu, and Pu]{zhu2024cosalpure}
Jiayi Zhu, Qing Guo, Felix Juefei-Xu, Yihao Huang, Yang Liu, and Geguang Pu.
\newblock Cosalpure: Learning concept from group images for robust co-saliency detection.
\newblock In \emph{Proceedings of the IEEE/CVF Conference on Computer Vision and Pattern Recognition}, pages 3669--3678, 2024.

\end{thebibliography}
}

\newpage
\appendix
\onecolumn

\section{Appendix}
\label{sec:appendix}

\paragraph{Experiment Setup}
Our 3D reconstruction and 2D inpainting method is implemented on a single RTX 4090. We use the default parameters of 3DGS for reconstruction, generating a reconstructed render every 10,000 iterations. Additionally, we employed the commonly used stable-diffusion-v1-5 \citep{rombach2022high} as the base inpainting model, training it for 3,000 iterations (taking approximately 1.5 hours) using textual inversion for scene representation. Our diffusion model inference consists of a 50-step denoising process, initialized with a noise strength of 1.0 that is progressively reduced by a factor of 0.2 at each iteration.

% \subsection{Discussions}

% This section will combine experimental results to discuss the reasons behind some hyperparameter settings in \Cref{sec:method}, further demonstrating our approach. More details can be found in \Cref{sec:appendix}.

%
%

%
\paragraph{Time cost analysis and comparison.} 
To quantitatively evaluate performance and computational efficiency, we compare our method against baseline approaches (InFusion, SPIn-NeRF, and SpotLess) on the synthetic scene shown in \Cref{fig:extreme_case}. This scene provides ground truth data, enabling evaluation through reference-based metrics for both rendering quality and computational efficiency during optimization.

\begin{table}[h]
    \centering
    \resizebox{0.4\linewidth}{!}{
     \begin{tabular}{l|ccc}
        \toprule
        Method & LPIPS $\downarrow$ & PSNR $\uparrow$ & Time Cost \\
        \midrule
        InFusion & 0.23 & 19.34 & 16m 34s\\
        SPIn-NeRF& 0.15 & 23.33 & 7h 32m 18s\\
        SpotLess & 0.14 & 24.75 & 30m 26s\\
        \midrule
        Ours & 0.10 & 26.38 & 33m 34s\\
        \bottomrule
    \end{tabular}
    }
    \caption{Quantitative results and time costs on the synthesis data.}
    \label{tab:timecost-quality-systhensis-data}
\end{table}

As shown in \Cref{tab:timecost-quality-systhensis-data}, while our method incurs additional computational overhead compared to vanilla 3DGS due to the integration of iterations and diffusion models, it achieves superior rendering quality while maintaining comparable efficiency to state-of-the-art 3DGS methods (e.g., SpotLess \citep{sabourgoli2024spotlesssplats}). Furthermore, our approach demonstrates significantly better reconstruction quality while being approximately 10× faster than leading NeRF-based methods such as SPIn-NeRF.

%
% \subsubsection{How many tokens do we need to depict one scene?}

\begin{figure}[h]
    \centering
    \includegraphics[width=0.99\linewidth]{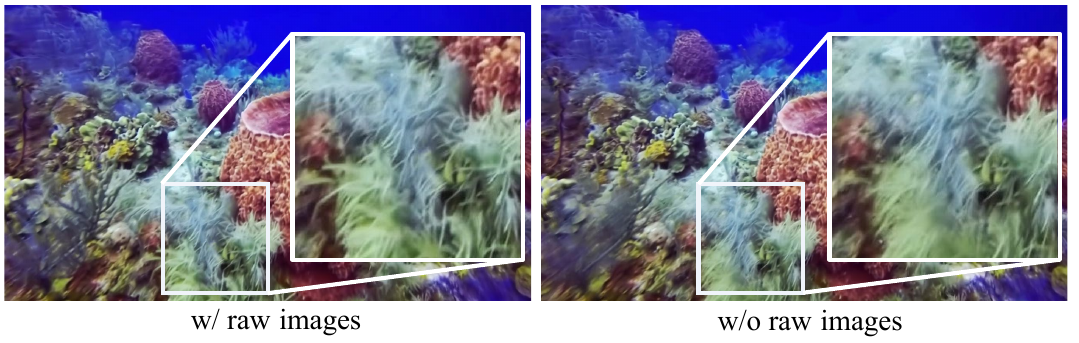}
    \caption{Reconstruction results with and without raw images. Involving the raw images in our method will improve the inpainting performance.} 
    \label{raw_image_ablation}
\end{figure}

\paragraph{Reasons for combining raw images $\mathcal{I}$ and $\tilde{\mathcal{I}}$ rather than substituting raw images $\mathcal{I}$ with $\tilde{\mathcal{I}}$ in \Cref{subsec:distrator}}
As shown in \Cref{raw_image_ablation}, the reconstruction without raw images could not render the seaweed without ambiguity. The accumulated error from two iterations, caused by 3DGS's inability to fit the scene fully and the uncertainty introduced by the generated model, deteriorates the image quality. Raw images act as an "anchor" for our method, ensuring that the rendered images align closely with the input images and do not deviate significantly.

% \subsubsection{How does the quality of priors infer the inpainting results? }
% By Yu: Please double check the writing of semantic information detection markers: "In contrast, static objects require the introduction of semantic information detection markers to be repaired."
\subsection{Impact of Prior (Mask) on Inpainting}
% \paragraph{Impact of Prior (Mask) on Inpainting}

Adding the prior (mask) information in our method will significantly improve the inpainting results, especially for those static objects. This is easy to understand because dynamic objects create inconsistencies during the reconstruction process, which our algorithm can detect. In contrast, static object inpainting necessitates the semantic information the detection model identifies. 

\begin{figure}[h]
\centering
\includegraphics[width=0.99\linewidth]{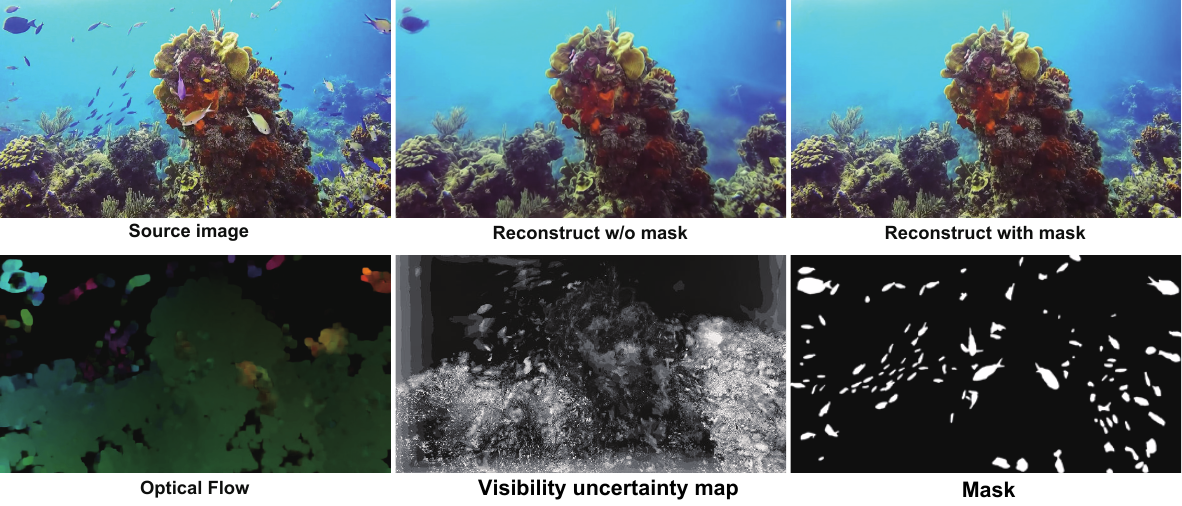}\\
\caption{Impact of prior (mask) on inpainting results. Our method will improve inpainting performance by incorporating the mask information. To analyze the results, we also display the optical flow of the source image and the visibility uncertainty map.}
\label{fig:mask_ablation}
\end{figure}

For instance, in the top-left corner of \Cref{fig:mask_ablation}, the fish is retained while the others are removed. This is primarily because the fish remains stationary across different views (as evident in the optical flow map of \Cref{fig:mask_ablation}, where the top-left fish exhibits low flow values at its center). Consequently, it has a lower value in the visibility uncertainty map (see the corresponding map in \Cref{fig:mask_ablation}). Without using a mask to label this area for repair manually, the fish's geometric characteristics resemble those of a stationary object, such as a rock, making it indistinguishable from our uncertainty detection system.

In contrast, moving fish create significant geometric inconsistencies across viewpoints, enabling our uncertainty detection to flag them as anomalies. This leads to their removal through the inpainting process.
To address these challenging scenarios, we introduced mask annotations for fish detection, providing semantic guidance for our inpainting method. As shown in the last column of \Cref{fig:mask_ablation}, incorporating the mask ensures the successful removal of the top-left fish.

\begin{figure}[h]
\centering
\includegraphics[width=0.99\linewidth]{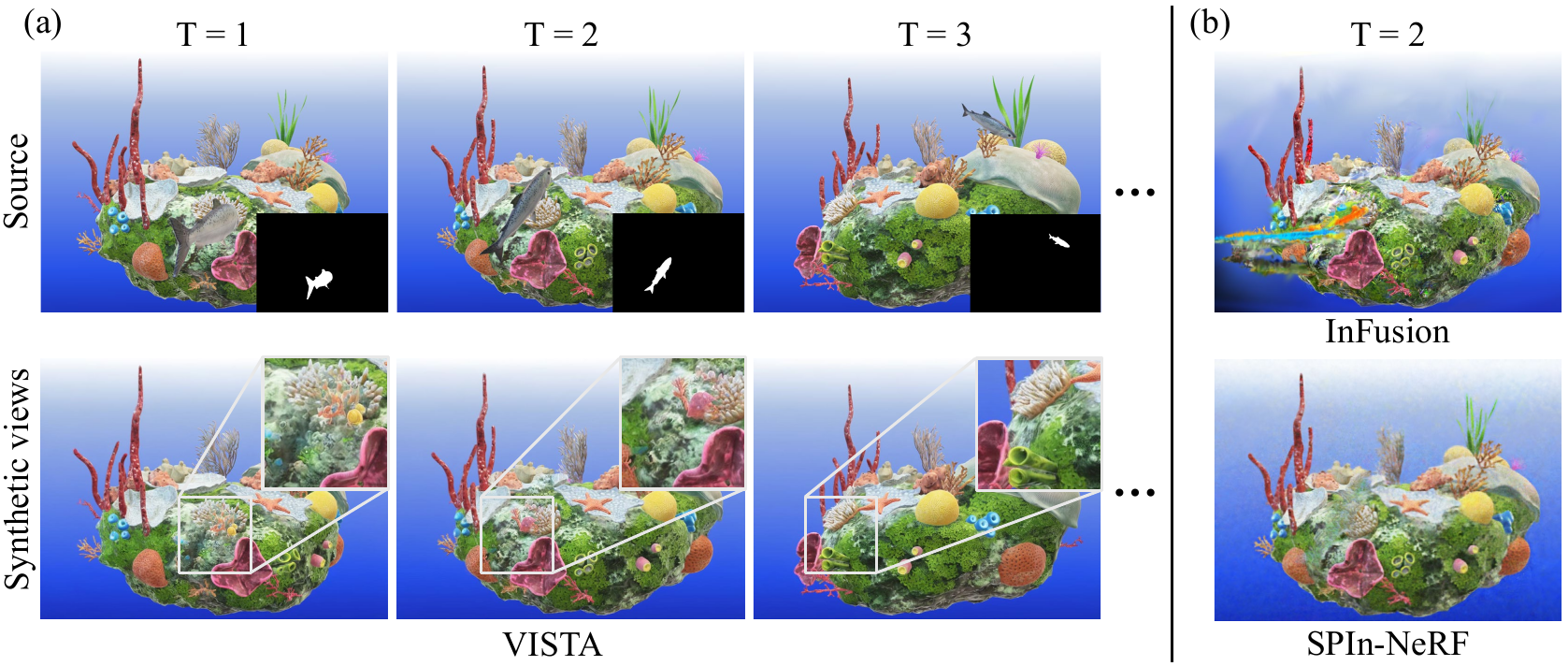}\\
\caption{\textbf{(a) The figure of an artificial synthesis scene in extreme cases.} The original views of three adjacent cameras and the inpainting results of our method are demonstrated for comparison. \textbf{(b) The results of InFusion and SPIn-NeRF in extreme cases.} Their results are obtained by the camera from `T = 2' in (a).} 
\label{fig:extreme_case}
\end{figure}

\paragraph{VISTA in limited scenarios} Our uncertainty maps are built by observing a set of adjacent perspectives/views, thus fully utilizing complementary visual cues. However, in some extreme conditions, we don't have enough valid adjacent perspectives/views to get the visual cues. To investigate the performance of our method under such extreme conditions, we manually synthesized an extreme scenario where the camera rapidly changes poses, resulting in very few available adjacent viewpoints. In this case, the VISTA-GI can hardly detect the inconsistency between different views, requiring VISTA-CL to produce better results.

As shown in \Cref{fig:extreme_case} (a), thanks to the 2D diffusion model, our method utilizes its results to effectively inpaint the scene in such extreme conditions. Meanwhile, as shown in \Cref{fig:extreme_case} (b), The InFusion result is unrealistic due to neglecting consistency in inpainting. SPIn-NeRF shows a reasonable result but with blurry and indistinguishable inpainting areas. Compared to other methods, our approach benefits from Scene Conceptual Learning, resulting in clearer and more reasonable repairs in the target areas, and the textures and content maintain consistency with the original scene.

\subsection{Visualization of our method}

In this part, we visualize more results to demonstrate the effectiveness of our method and the potential failure scenarios that may arise.

\begin{figure}[h]
\centering
\includegraphics[width=0.95
\linewidth]{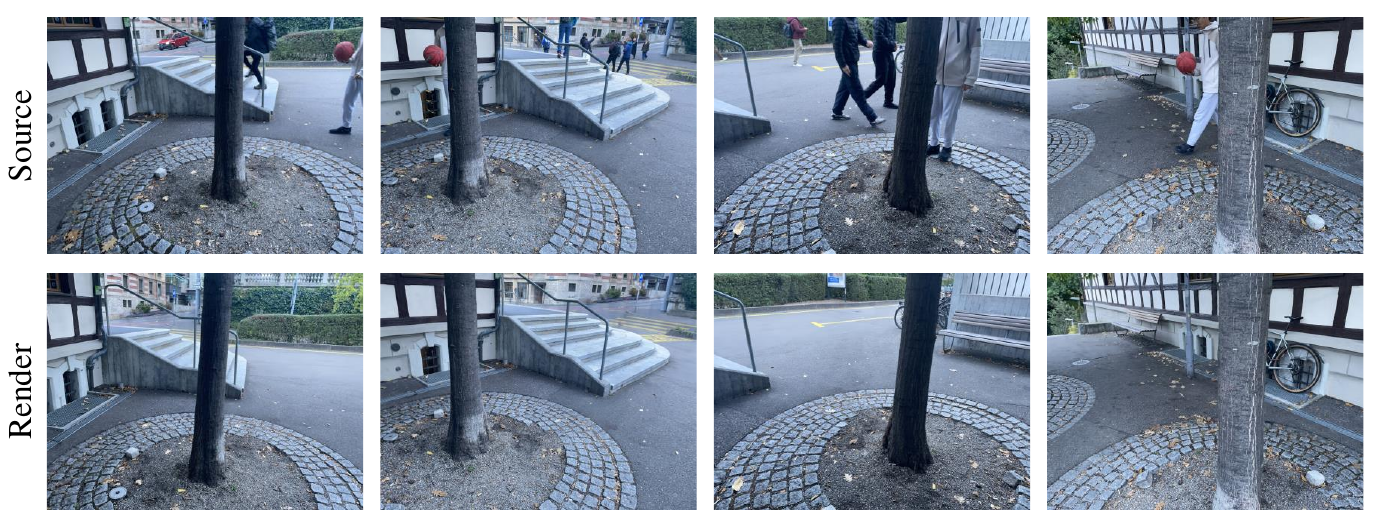}
\caption{Case of real-world pedestrian removal from the nerf-on-the-go dataset.} 
\label{fig:realworldcase}
\end{figure}

\paragraph{Real-world case.} The underwater dataset we used is derived from real-world diving videos, and due to the effects of the underwater medium and floating debris, these scenes are challenging scenarios in the real world. We also tested our dataset on a scene related to pedestrian removal from the nerf-on-the-go dataset. This scene, called Tree, contains 212 images, with the main distractors from moving pedestrians. As shown in \Cref{fig:realworldcase}, our method achieved high-quality results on this dataset. Due to the abundance of viewpoints in the dataset, there is a lot of supplementary information between perspectives, allowing our method to effectively utilize other viewpoints to repair the blurring caused by moving distractors.

\begin{figure}[h]
\centering
\includegraphics[width=0.95
\linewidth]{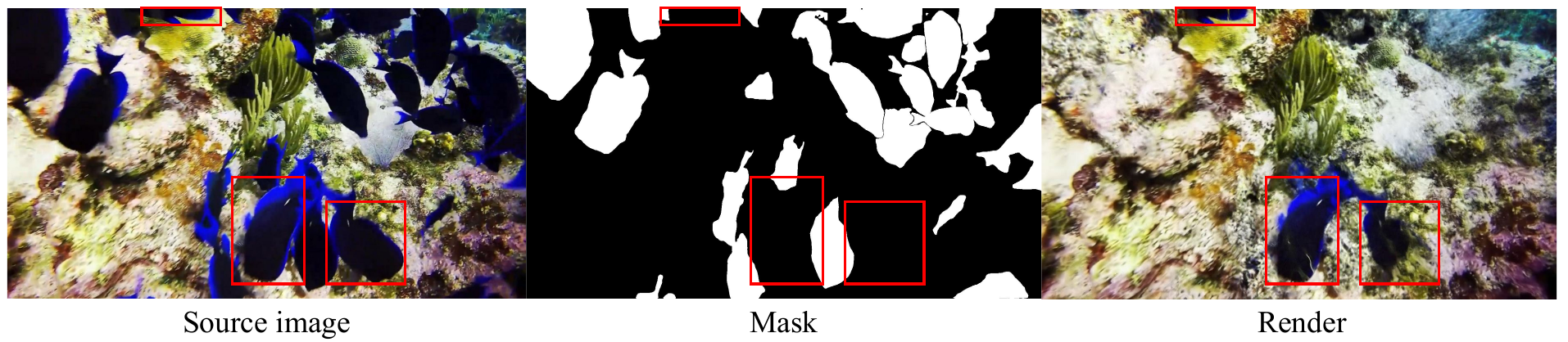}
\caption{Failure case from our underwater 3D inpainting dataset.} 
\label{fig:failcase}
\end{figure}

\paragraph{Fail case from our dataset.} We provide failure cases of our algorithm in \Cref{fig:failcase}. Due to errors in the prior mask, some fish were not detected by the object detection model. Furthermore, since the fish did not move significantly during the shooting process, these areas did not produce inconsistencies across multiple viewpoints during reconstruction, making it difficult for the VISTA-GI component to identify these areas through uncertainty. This also validates our algorithm design approach: VISTA-CL introduces semantic information through masks, while VISTA-GI incorporates geometric information through uncertainty, complementing each other to remove distractors. However, in this failed case, issues arose in both aspects, resulting in poor reconstruction quality of the final scene.

\begin{figure}[h]
\centering
\includegraphics[width=1.0
\linewidth]{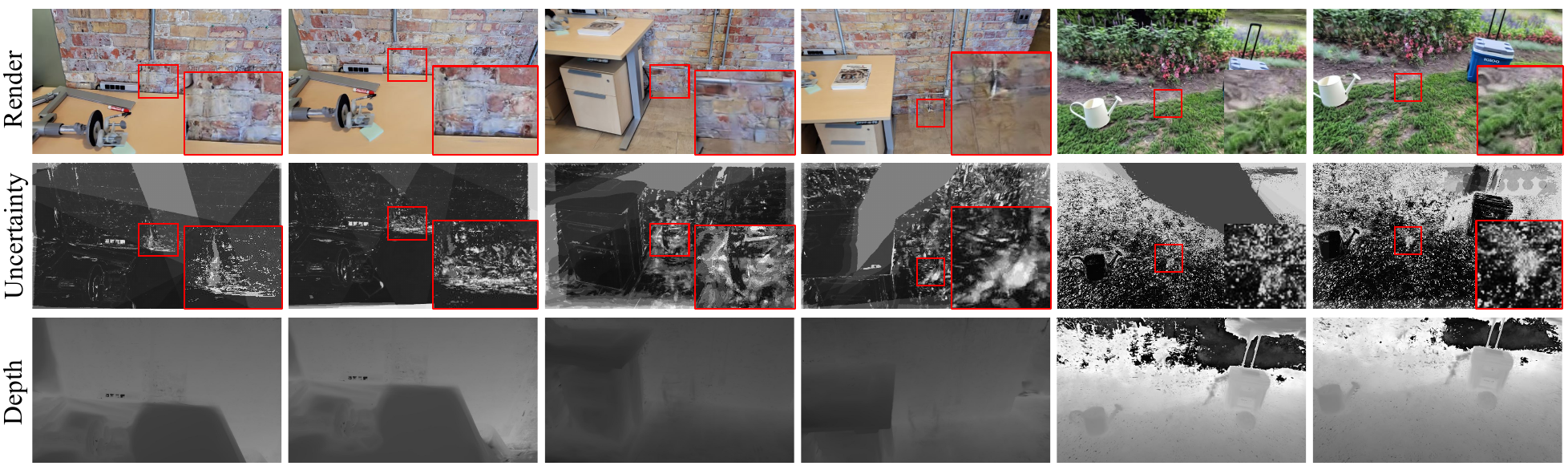}
\caption{Visualization of the uncertainty map and depth of static scenes.} 
\label{fig:static_uncertainty_map}
\end{figure}

\paragraph{Uncertainty and depth maps of static scenes.} As shown in \Cref{fig:static_uncertainty_map}, we further visualize the uncertainty and depth maps of the static scenes. The deeper the color, the closer the depth. It can be observed that our method identifies areas in the rendered image that are inconsistent with other viewpoints and generates reasonable contents.

% \subsection{Impacts of Hyper-parameters}

% In this part, we study the influence of the hyper-parameter $\vartheta$ in \reqref{eq:updated_mask}, the initial noise \& iterations of diffusion inference, and the threshold of the uncertainty map.

\begin{table}[hbt]
    \centering
    \resizebox{0.4\linewidth}{!}{
     \begin{tabular}{c|ccc}
        \toprule
        Resolution& LPIPS $\downarrow$& PSNR $\uparrow$&SSIM $\uparrow$\\
        \midrule
        64$\times$64& 0.51& 16.27&0.68\\
        128$\times$128& 0.42& 18.89&0.69\\
        256$\times$256& 0.26& 21.33&0.71\\
        512$\times$512& 0.11& 26.04&0.84\\
        \midrule
        1299$\times$974 & 0.10& 26.38&0.86\\
        \bottomrule
    \end{tabular}
    }
    \caption{Quantitative ablation results of different resolutions. \label{tab:resolution_ablation}}
\end{table}

\subsection{Impacts of Different Image Resolution}

% \revised{In our experiment setup, we use the stable diffusion v1.5 as the inpainting model and train and test the model following its default setup: if the input image has a resolution higher than 512$\times$512, we crop the image to a new size that is both the closest to the original image size and a multiple of 8; if the input image is smaller than 512x512, we rescale the image to 512$\times$512. To analyze the influence of the strategy on different original resolutions, given an original scene with input images having the size of 1299$\times$974, we downsample these images to four resolutions: 64$\times$64, 128$\times$128, 256$\times$256, 512$\times$512. Then, as shown in \Cref{tab:resolution_ablation}, we can build a 3D model and evaluate the rendering quality for each resolution.}

In our experiment setup, we use the stable diffusion v1.5 as the inpainting model and train and test the model following its default setup: if the input image has a resolution higher than 512$\times$512, we crop the image to a new size that is both the closest to the original image size and a multiple of 8; if the input image is smaller than 512$\times$512, we rescale the image to 512$\times$512. To analyze the influence of the strategy on different original resolutions, given an original scene with input images having a size of 1299$\times$974, we downsample these images to four resolutions: 64$\times$64, 128$\times$128, 256$\times$256, and 512$\times$512. Then, for each resolution, we can build a 3D model and evaluate the rendering quality. As shown in \Cref{tab:resolution_ablation}, we observe that: (1) reducing the resolution to 512$\times$512 does not significantly impact any of the metrics, demonstrating our method's robustness to substantial resolution changes. (2) further decreasing the resolution leads to gradual degradation in reference-based metrics, while non-reference metrics remain relatively stable.

% \subsection{Quantitative Analysis of Large Viewpoint Differences}

\end{document}